\documentclass[journal]{IEEEtran}
\usepackage[algpseudocode=false]{signalpreamble}
\usepackage{algorithmic}
\usepackage{algorithm}
\usepackage{tikz}
\usepackage{pifont}
\usetikzlibrary{shapes.misc}
\usetikzlibrary{arrows.meta, positioning, calc}
\usetikzlibrary{shapes}

\newacronym[plural=GNNs,firstplural=Graph Neural Networks (GNNs)]{GNN}{GNN}{graph neural network}
\newacronym[plural=FDEs,firstplural=fractional-order differential equations (FDEs)]{FDE}{FDE}{fractional-order differential equation}
\newacronym{MSE}{MSE}{mean squared error}
\newacronym{PDE}{PDE}{partial differential equation}
\usepackage[textsize=tiny]{todonotes}

\begin{document}
\title{Adaptive Multi-view Graph Contrastive Learning via Fractional-order Neural Diffusion Networks}
\author{Yanan Zhao, Feng Ji, Jingyang Dai, Jiaze Ma, Keyue Jiang, Kai Zhao, Wee Peng Tay
\thanks{Yanan Zhao, Feng Ji, Jingyang Dai, Jiaze Ma, Kai Zhao, and Wee Peng Tay are with the School of Electrical and Electronic Engineering, Nanyang Technological University, Singapore.}
\thanks{Keyue Jiang is with University College London, London, U.K.}
}


\maketitle

\begin{abstract}
Graph contrastive learning (GCL) learns node and graph representations by contrasting multiple views of the same graph. Existing methods typically rely on fixed, handcrafted views-usually a local and a global perspective—which limits their ability to capture multi-scale structural patterns. We present an augmentation-free, multi-view GCL framework grounded in fractional-order continuous dynamics. By varying the fractional derivative order $\alpha\in(0,1]$, our encoders produce a continuous spectrum of views: small $\alpha$ yields localized features, while large $\alpha$ induces broader, global aggregation. We treat $\alpha$ as a learnable parameter so the model can adapt diffusion scales to the data and automatically discover informative views. This principled approach generates diverse, complementary representations without manual augmentations. Extensive experiments on standard benchmarks demonstrate that our method produces more robust and expressive embeddings and outperforms state-of-the-art GCL baselines.
\end{abstract}


\section{Introduction}\label{sec:int}

Contrastive learning is a self-supervised paradigm that trains representations by pulling semantically similar pairs together and pushing dissimilar pairs apart in the embedding space. It uncovers useful structure without labels and has produced strong results in computer vision, natural language processing, and recommendation systems \cite{Radford2021,Grill2021,Tchen2020}. Applied to graph data, this paradigm is known as \emph{Graph Contrastive Learning (GCL)}, where positive pairs are typically derived from different views (e.g., node-, subgraph-, or graph-level perturbations) of the same instance and negatives from other, unrelated instances.

A central challenge in GCL is how to produce distinct, semantically meaningful views. Prior approaches follow two broad strategies. Augmentation-based methods synthesize multiple views by perturbing the graph (e.g., edge deletion, feature masking), while augmentation-free methods obtain different views by applying distinct encoders to the same input. Despite these differences, most methods still rely on a small set of fixed, handcrafted views—commonly a ``local'' and a ``global'' perspective—which constrains their ability to represent multi-scale structure. For example, BGRL \cite{thakoor2022BGRL} creates two views via structural and feature perturbations, and PolyGCL \cite{chen2024polygcl} constructs low-pass and high-pass views through spectral filtering. Such designs increase view diversity but remain rigid and often require manual tuning. This motivates two questions:
\begin{enumerate}[align=left,leftmargin=10pt,label=\textbf{Q\arabic*}]
    \item\label{Q1} \textbf{How can we adaptively generate a diverse set of views that capture multi-scale semantics in GCL, moving beyond fixed, handcrafted local and global perspectives?}
    \item\label{Q2} \textbf{Can this be achieved without relying on heuristic augmentations?}
\end{enumerate}

To answer these questions, we propose a novel \textbf{multi-view augmentation-free} framework grounded in fractional-order continuous dynamics. Our key insight is to leverage \emph{\glspl{FDE}} \cite{Sivalingam2025} to model graph diffusion and generate a continuum of semantically diverse views, each parameterized by a distinct fractional derivative order. Prior works, which rely on ordinary differential equations (ODEs) \cite{chamberlain2021grand, thorpe2022grand++, chamberlain2021blend, SonKanWan:C22, rusch2022graph, KanZhaSon:C23, ZhaKanSon:C23b},  simulate graph diffusion as continuous-time heat flow and serve as smooth analogues of message passing in models like GCN. Unlike such ODE-based approaches, FDEs model graph diffusion by introducing the fractional derivative operator \(\frac{\ud^\alpha}{\ud t^\alpha}\), where $\alpha\in (0,1]$ \cite{ZhaKanJi:C24, KanZhaDin:C24frond, ZhaKanSon:C24robustfrond, KanLiZha25, CuiKanLi25, CuiLiLi25}. When $\alpha=1$, the standard ODE is recovered; however, for $\alpha < 1$, the operator incorporates non-local information through memory effects, allowing the diffusion to reflect long-range temporal dependencies and attenuated propagation. 

By leveraging FDEs, {\bf multi-view} GCL emerges naturally: the fractional order $\alpha\in(0,1]$ provides a continuous control of diffusion scale. Varying $\alpha$ yields a continuum of views — small $\alpha$ emphasizes local components, while large $\alpha$ produces broader global aggregation. This gives a principled, augmentation‑free mechanism to generate diverse and complementary views without handcrafted filters (e.g., filter‑based methods such as PolyGCL) or task‑specific perturbations (e.g., BGRL). Moreover, $\alpha$ is a learnable continuous parameter, so the model can automatically adapt diffusion scales to dataset heterogeneity. To our knowledge, this is the first work to exploit the multi‑scale behaviour of fractional‑order dynamics for contrastive learning. Our main contributions are summarized as follows:
\begin{itemize}
\item We propose a novel multi-view, augmentation-free GCL framework based on neural diffusion encoders governed by \emph{fractional-order dynamics}. This approach is conceptually simple, computationally efficient, and highly flexible, with each feature view controlled by a continuous parameter $\alpha \in (0,1]$, naturally spanning a spectrum of diffusion scales.

\item Our main theoretical result shows that embeddings produced by FDE encoders at different fractional orders are provably distinct, and that the separation between these views grows as the gap between the orders $\alpha$ increases. To our knowledge, this is the \textbf{first} formal analysis of this multi-scale distinguishability in the context of contrastive learning, despite prior empirical use of FDEs in GNNs.

\item We tackle two core challenges in GCL: (i) dimension collapse — mitigated by including encoders with small fractional orders that yield less energy‑concentrated, higher‑rank embeddings; and (ii) view collapse — prevented via a regularized contrastive objective that penalizes alignment of dominant directions across views, promoting diverse, complementary representations without requiring negative samples.

\item The fractional orders $\alpha$ are treated as learnable parameters, allowing the model to discover informative diffusion scales from data instead of relying on manual tuning or grid search. To control complexity, we propose a view‑learning strategy that prunes redundant scales while retaining diverse, complementary views.

\item We present a rigorous stability analysis, deriving explicit perturbation bounds that quantify how input, parameter, and topological disturbances affect encoder outputs. Empirically, our approach consistently outperforms state‑of‑the‑art robust GCL methods under both black‑box and white‑box attack scenarios.

\item We evaluate on a broad set of homophilic and heterophilic graph benchmarks. Empirical results demonstrate that our method consistently outperforms state‑of‑the‑art GCL baselines, confirming the effectiveness of the proposed multi‑scale view generation mechanism.
\end{itemize}

\section{Related work}

GCL methods can be categorized into augmentation-based and augmentation-free approaches. Augmentation-based methods synthesize multiple views by perturbing the input graph (e.g., edge deletion, feature masking) \cite{YifeiCGKS2023, zhu2020GRACE, zhu2021GCA, Teng2022DSSL, Chen2022HGRL,XuEPAGCL2025}. Augmentation-free methods instead obtain different views by applying distinct encoders or filters to the same graph \cite{peng2020gmi, zhang2022LGCL}. Many augmentation-free approaches (e.g., GraphACL \cite{Teng2023GraphACL}, SP-GCL \cite{Wang2023SPGCL}, GraphECL \cite{xiaographecl2024}, PolyGCL \cite{chen2024polygcl}, LOHA \cite{Loha2025}, AFECL \cite{LiAFECL2025}) still depend on negative samples for contrastive training. SDMG \cite{zhu2025sdmg} follows a diffusion-based, self-supervised design using two low-frequency encoders to emphasize global signals. By contrast, methods such as BGRL \cite{thakoor2022BGRL} and CCA-SSG \cite{Zhang2021CCA_SSG} remove negative sampling but keep augmentations, while AFGRL \cite{Lee2022AFGRL} removes both negatives and augmentations yet can suffer from limited view diversity and fixed propagation depth, which hurts performance on heterophilic graphs.

\begin{table}[tb]
\caption{FD-MVGCL (ours) vs. SOTA GCL Methods.}
\centering
\fontsize{8pt}{9pt}\selectfont
\setlength{\tabcolsep}{1.5pt}
\begin{tabular}{lccccc}
\toprule
Method            & $>2$ views   & Aug-Free & Neg-Free & Hetero. & Adapt.\\
\midrule
BGRL ('22)        & \ding{55}    & \ding{55} & \ding{51} & \ding{51} & \ding{51} \\
AFGRL ('22)       & \ding{55}    & \ding{51} & \ding{51} & \ding{55} & \ding{55} \\
GraphACL ('23)    & \ding{55}    & \ding{51} & \ding{55} & \ding{51} & \ding{55} \\
GraphECL ('24)    & \ding{55}    & \ding{51} & \ding{55} & \ding{51} & \ding{55} \\
PolyGCL ('24)     & \ding{55}    & \ding{51} & \ding{55} & \ding{51} & \ding{55} \\
LOHA ('25)        & \ding{55}    & \ding{51} & \ding{55} & \ding{51} & \ding{55} \\
AFECL ('25)       &\ding{55}     & \ding{51} & \ding{55} & \ding{55} & \ding{55} \\
EPAGCL ('25)       &\ding{55}     & \ding{55} & \ding{55} & \ding{55} & \ding{51} \\
\midrule
FD-MVGCL   &\ding{51}     & \ding{51} & \ding{51} & \ding{51} & \ding{51} \\
\bottomrule
\end{tabular}
\label{tab:methods_comparison}
\end{table}

In contrast, our method operates in an augmentation‑free and negative‑sample‑free manner and performs well on both homophilic and heterophilic graphs (see \cref{tab:methods_comparison}). We propose a different multi‑view mechanism based on fractional‑order neural diffusion: each view is generated by an FDE with its own fractional order $\alpha\in(0,1]$, providing continuous, learnable control over the diffusion scale. This yields diverse, complementary views without handcrafted filters or graph augmentations. By learning the $\alpha$ values from data, the model removes the need for manual tuning or grid search. To our knowledge, this is the first work to generate multi‑scale contrastive views via fractional‑order dynamics on graphs, and it achieves strong results across both homophilic and heterophilic settings.

\section{Preliminaries}

\subsection{Problem formulation}

Consider an undirected graph $\calG=(\calV,\calE)$, where $\calV=\{v_1,\ldots, v_N\}$ is a finite set of $N$ nodes and $\calE \subset \calV \times \calV$ is the set of edges. The raw features of the nodes are represented by the matrix $\bX$, with the $i$-th \emph{row} corresponding to the feature vector $\bx_i$ of node $v_i$. The weighted symmetric adjacency matrix $\bA = (a_{ij})$ has size $N \times N$, where $a_{ij}$ denotes the edge weight between nodes $v_i$ and $v_j$. We denote the complete graph information by $\calX = (\bA, \bX)$. 

In unsupervised feature learning, an encoder $f_\theta$ maps each raw feature $\bx_i$ to a latent representation $\by_i = f_\theta(\calX, v_i) \in \bbR^F$, forming the embedding matrix $\bY \in \bbR^{N \times F}$ for downstream tasks such as node classification. Multi-view GCL introduces self-supervision by encouraging consistency among node embeddings $\bz_i$ produced by multiple encoders $f_{\theta_k}$, $k \geq 2$.

\subsection{Graph neural diffusion models: from GCN to FDE}

Traditional \gls{GNN} models, such as GCN and GAT \cite{kipf2017semi, velickovic2018graph, hamilton2017inductive}, rely on (discrete) graph message passing for feature aggregation. In the $k$-th iteration, the message passing step is given by $\bY\tc{k}=\overline{\bA}\bY^{(k-1)}$, where $\overline{\bA}$ is the normalized adjacency matrix of $\bA$, and $\bY^{(k-1)}$ is the output from the $(k-1)$-th iteration. In contrast, \cite{chamberlain2021grand} introduces a continuous formulation based on heat diffusion, where the feature evolution $\bY(t)$ is governed by an ordinary differential equation (ODE):
\begin{align}
\label{eq.ode}
    \ddfrac{}{t} \bY(t) = \calF(\bW,\bY(t))
\end{align}
with the initial condition $\bY(0)$ set to $\bX$ or its transformation. Here, $t$ serves a role analogous to the layer index in GCNs, and $\calF(\bW,\bY(t))$ defines the spatial diffusion. Let $\overline{\bL} = \bI - \overline{\bA}$ denote the \emph{normalized Laplacian}. The GRAND model \cite{chamberlain2021grand} is given by $\calF(\bW,\bY(t)) = -\overline{\bL}\bY(t)$.

This formulation is further extended in \cite{KanZhaDin:C24frond} by incorporating \glspl{FDE}. Specifically, for each \emph{order parameter} $\alpha \in (0,1)$, the corresponding fractional-order differential operator $D^{\alpha}_t$ is defined as 
\begin{align*}
     D^{\alpha}_tf(t) = \frac{1}{\Gamma(-\alpha)}\int_0^{\infty} \frac{f(t-\tau)-f(t)}{\tau^{1+\alpha}}\ud\tau, 
\end{align*}
where $\Gamma(\cdot)$ denotes the Gamma function.

The operator $D^{\alpha}_t$ extends the concept of $\ddfrac{}{t}$ (from \cref{eq.ode}) such that as $\alpha \to 1$, $D^{\alpha}_t$ converges to $\ddfrac{}{t}$. Thus, it is natural to define $D^1_t = \ddfrac{}{t}$, allowing $\alpha$ to be selected from the interval $(0,1]$. By generalizing \cref{eq.ode}, we define the dynamics of the features $\bY(t)$ under an FDE as:  
\begin{align}
D^\alpha_{t} \bY(t) = \calF(\bW,\bY(t)). \label{eq.frond_main}
\end{align}
The precise definition of $D^{\alpha}_t$ may vary slightly depending on the context (see \cref{appendix.fde_models}), but all formulations share a common characteristic: for $\alpha \in (0,1)$, $D^{\alpha}_t$ is defined via an integral. This implies that $\bY(t)$ depends on the entire history of $\bY(t')$ for $t' < t$. Consequently, unlike the solution to \cref{eq.ode}, the dynamics of $\bY(t)$ exhibit a ``memory effect.'' 

Our \emph{main insight} for GCL lies in leveraging the fractional order parameter $\alpha$ to systematically generate a diverse set of distinct encoders, enabling augmentation-free GCL. The multi-view framework is naturally facilitated by the continuous and parameterizable nature of $\alpha$, which allows for seamless integration and optimization through gradient descent. Our theoretical analysis focuses on the F-GRAND-l model \cite{KanZhaDin:C24frond} given by
\begin{align}
D^\alpha_{t} \bY(t) = -\overline{\bL}\bY(t). \label{eq.F-GRAND-l}
\end{align}
The analysis shall be performed within the framework of graph signal processing (GSP), which is reviewed next.

\subsection{Graph Signal Processing} GSP \cite{Shu13} provides a useful framework to analyze the impact of different $\alpha$ values on the generated views. We present a concise overview here. Since the normalized Laplacian $\overline{\bL}$ is symmetric, it admits an orthogonal decomposition $\overline{\bL} = \bU \bLambda \bU\T$. Here, the diagonal entries $\lambda_1 \leq \dots \leq \lambda_N$ of $\bLambda$ are the eigenvalues of $\overline{\bL}$, also known as the \emph{graph frequencies}, and the $i$-th column $\bu_i$ of $\bU$ is the eigenvector corresponding to $\lambda_i$. Each $\bu_i$ represents a signal on the graph $\calG$. For small $\lambda_i$, $\bu_i$ is smooth, meaning the signal values vary gradually across the edges. Conversely, for large $\lambda_i$, $\bu_i$ tends to be spiky, emphasizing local variations.

Any signal $\bx$ (e.g., a column of the feature matrix $\bX$) can be expressed as a \emph{spectral decomposition}:
\begin{align*}
\bx = \sum_{i=1}^{N} c_i\bu_i, \quad \text{where } c_i = \ip{\bx}{\bu_i}.
\end{align*}
In GSP, the coefficients $c_i$ are referred to as \emph{Fourier coefficients}, which quantify the response of $\bx$ to the basis vector $\bu_i$. A signal $\bx$ is said to have \emph{large smooth components} if $|c_i|$ is relatively large for small indices $i$, and it is considered \emph{energy concentrated} if $|c_i|$ is small for most indices $i$.

\section{The proposed model: FD-MVGCL}\label{CL_framework}

In this section, we introduce the adaptive \textbf{F}ractional \textbf{D}iffusion-based \textbf{M}ulti-view \textbf{G}raph \textbf{C}ontrastive \textbf{L}earning model (FD-MVGCL). We begin by presenting the multi-view generation mechanism inspired by \glspl{FDE}, supported by rigorous theoretical analysis and numerical validations. Next, we describe the FD-MVGCL architecture and address two key challenges in graph contrastive learning: dimension collapse and view collapse. Additionally, we propose an adaptive algorithm that automatically determines the optimal number of views and their corresponding fractional orders. Finally, we provide a stability analysis under graph perturbations, highlighting the robustness of FD-MVGCL.

\subsection{Contrasting encoders}
\subsubsection{Multi-view generation} \label{sec:dwd}

We adopt a non‑Markovian random‑walk interpretation \cite{KanZhaDin:C24frond} to illustrate how varying the fractional order $\alpha$ produces multi‑scale views. Under the FDE model \cref{eq.F-GRAND-l}, let ${}_{i}\bP(t)\in\mathbb{R}^N$ be the probability distribution of a fractional-order random walker over the node set $\calV$ when initialized at $v_i$; its $j$‑th entry is the probability of finding the walker at node $v_j$ at time $t$. The node embeddings are recovered as $\bY(t)=\sum_i {}_{i}\bP(t) \bY(0)_i$, where $\bY(0)_i$ is the initial feature (row) vector of $v_i$.
For $\alpha \in (0,1]$, the random walk on the graph $\calG$ evolves in infinitesimal time intervals $\Delta \tau > 0$ according to the following dynamics:
\begin{enumerate}[topsep=0pt, itemsep=2pt, partopsep=0pt, parsep=0pt, leftmargin=10pt]
    \item The walker remains at the current node for a random duration sampled from a heavy-tailed distribution $\psi_{\alpha}(n) = d_{\alpha} n^{-(1+\alpha)}$, where $d_{\alpha}>0$ ensures normalization. Smaller values of $\alpha$ result in heavier tails, leading to longer expected waiting times and a stronger memory effect.
    \item After waiting, the walker either transitions from node $i$ to a neighboring node $j$ with probability $(\Delta\tau)^{\alpha} d_{\alpha} |\Gamma(-\alpha)|\frac{W_{ij}}{d_i}$ for $i \neq j$, or remains at node $i$ with probability $1-(\Delta\tau)^{\alpha} d_{\alpha} |\Gamma(-\alpha)|$.
\end{enumerate}

This formulation provides an intuitive understanding of how the fractional parameter $\alpha$ governs the scale of information propagation over the graph:
\begin{itemize}[topsep=1pt, itemsep=2pt, partopsep=0pt, parsep=0pt, leftmargin=10pt]
    \item \textbf{Local view}: As $\alpha \to 0$, the walker exhibits long waiting times and infrequent transitions, resulting in highly localized behavior and aggregating information primarily from nearby nodes.
    \item \textbf{Global view}: As $\alpha \to 1$, the walker transitions more frequently and explores a wider portion of the graph, enabling global aggregation and capturing long-range dependencies.
\end{itemize}

To further elucidate how multi-scale views naturally arise from fractional-order derivatives, we provide a mathematical analysis of the distinct representations generated by FDE encoders with varying fractional orders $\alpha$. For clarity and conciseness, we present an informal summary of the results here, while a \emph{detailed and rigorous} explanation is deferred to \cref{sec:td}. To the best of our knowledge, this is the \emph{first} formal mathematical analysis of this phenomenon in the context of contrastive learning.

\begin{figure*}[!tb]
\centering


\includegraphics[width=1.01\linewidth]{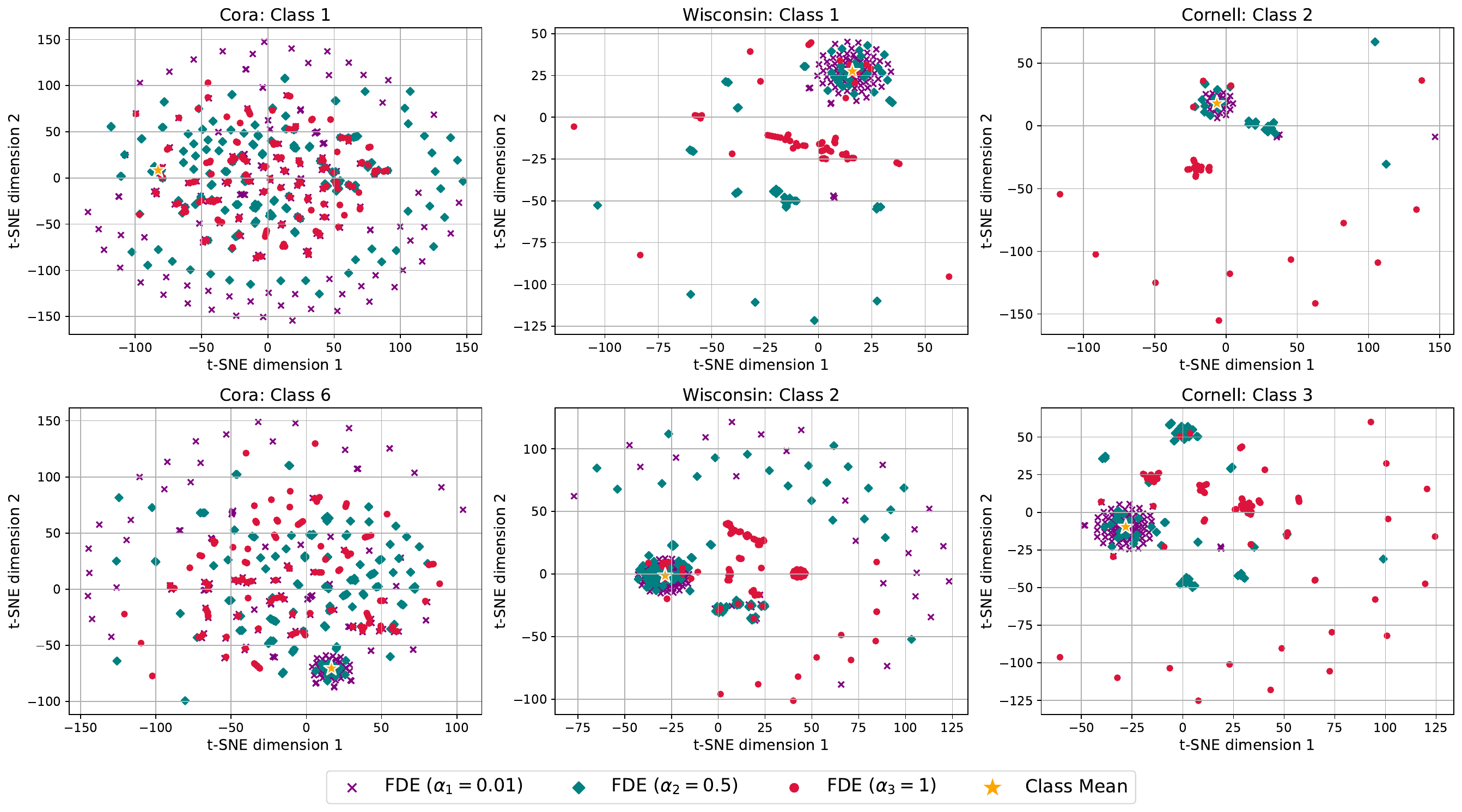}
\vspace{-6mm}
\caption{t-SNE visualizations of single-class node embeddings from encoders with different FDE orders are shown. Class means are aligned for fair comparison. Results on Cora (homophilic) and Wisconsin and Cornell (heterophilic) reveal distinct embedding behaviors: smaller $\alpha$ values yield compact, core-concentrated clusters, whereas larger $\alpha$ values produce more evenly dispersed feature distributions. Additional visualizations for other classes are provided in \cref{supp.tsne}.} \label{fig:wst}
\end{figure*}

\begin{Theorem}[Informal; see \cref{thm:sgi} in \cref{sec.td_1}] \label{thm:sgi-informal}
Assuming the model \cref{eq.F-GRAND-l}, for $0 < \alpha_l < \alpha_g \leq 1$, the following hold for features $\bY_{\alpha_l}(t)$ and $\bY_{\alpha_g}(t)$ when $t$ is large:
\begin{enumerate}[(a)]
    \item \label{it:bhl} $\bY_{\alpha_g}(t)$ contains more large smooth components as compared with $\bY_{\alpha_l}(t)$.
    \item \label{it:bim} $\bY_{\alpha_l}(t)$ is less energy concentrated as compared with $\bY_{\alpha_g}(t)$.
\end{enumerate}
Moreover, the contrast in (a) and (b) becomes more pronounced as the difference $\alpha_g - \alpha_l$ increases.
\end{Theorem}

As discussed in \cref{sec:int}, the effectiveness of augmentation-free GCL relies on the ability of different encoders to generate feature representations with distinct views. We argue that $\bY_{\alpha_g}(t)$ emphasizes \emph{global} structural patterns as $\alpha_g \to 1$, while $\bY_{\alpha_l}(t)$ captures finer \emph{local} details as $\alpha_l \to 0$. Specifically, by \cref{thm:sgi-informal}\ref{it:bhl}, the smooth components (associated with small graph frequencies $\lambda_i$) dominate in $\bY_{\alpha_g}(t)$. Each such component is spanned by a vector $\bu_i$, which represents a signal with minimal variation across edges, thereby encoding global structural information. Conversely, in $\bY_{\alpha_l}(t)$, non-smooth components $\bu_i$ (corresponding to large graph frequencies $\lambda_i$) have higher coefficients and thus contribute more significantly. Since these $\bu_i$ are localized (spiky), $\bY_{\alpha_l}(t)$ effectively captures local details. In summary, we establish the correspondence: $\alpha_l \longleftrightarrow$ ``local'' and $\alpha_g \longleftrightarrow$ ``global.''

By selecting distinct $\alpha$ values, we construct a set of FDEs, each capturing a unique propagation scale. This approach naturally generates diverse, multi-scale graph views (i.e., embeddings $\bY_{\alpha}(t)$), enriching graph contrastive learning. Numerical experiments demonstrate that embeddings $\bY_{\alpha_i}(t)$ and $\bY_{\alpha_j}(t)$ exhibit distinct patterns when $\alpha_i \neq \alpha_j$, as illustrated in \cref{fig:wst} for the Cora, Wisconsin, and Cornell datasets. For small $\alpha$ values (e.g., 0.01), features are tightly clustered near the class mean, while larger $\alpha$ values (e.g., 1) produce more dispersed feature distributions. Intermediate $\alpha$ values (e.g., 0.5) yield distributions that balance these extremes. This demonstrates that each FDE provides a semantically distinct and complementary view, enhancing the effectiveness of contrastive learning.

\subsubsection{Dimension collapse} \label{sec:ce}

Dimension collapse occurs when features are confined to a low-dimensional subspace within the full embedding space \cite{zbontarbarlowtwin2021}, which is a critical issue to address in contrastive learning. We argue that $\bY_{\alpha_l}(t)$, generated with a small $\alpha_l$, effectively mitigates this problem. As established in \cref{thm:sgi-informal}\ref{it:bim}, $\bY_{\alpha_l}(t)$ exhibits lower energy concentration compared to $\bY_{\alpha_g}(t)$ when $\alpha_l < \alpha_g$. This means that the columns of $\bY_{\alpha_l}(t)$ can be expressed as $\sum_{1\leq i\leq N} c_i\bu_i$, with a relatively larger number of significant $|c_i|$ values. As a result, the features are distributed across a higher-dimensional space, avoiding collapse into a low-dimensional subspace.

\begin{figure}[!htb]
\centering
\includegraphics[width=0.95\columnwidth, trim={3cm 10.8cm 3cm 9.6cm},
        clip]{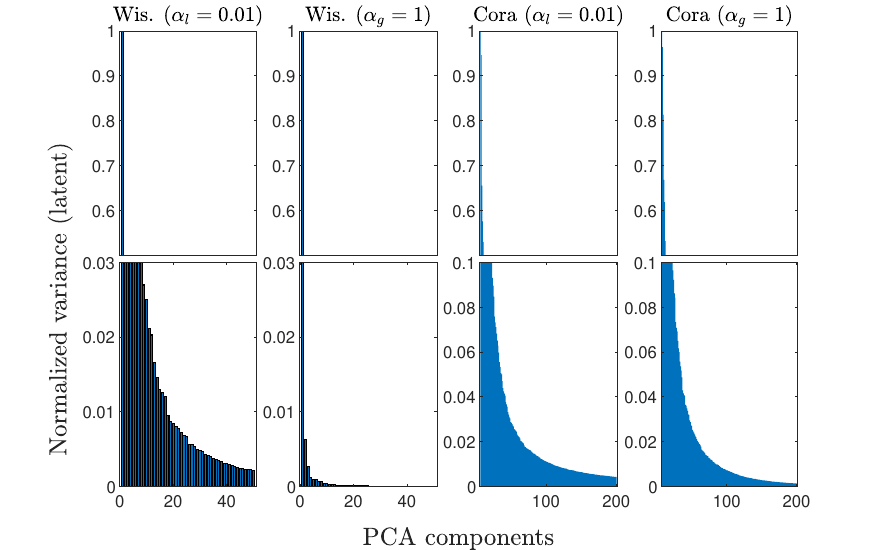}
\caption{The PCA components of features for different datasets and choices of FDE order parameters. We see that for at least the small order $\alpha_l$, the bar chart is more spread out, which prevents dimension collapse.} \label{fig:wsp}
\end{figure}

To support this claim, we perform principal component analysis (PCA) on $\bY_{\alpha_l}(t)$ and $\bY_{\alpha_g}(t)$ for the Cora and Wisconsin datasets. The results, shown in \cref{fig:wsp}, confirm our assertions.

\subsubsection{Encoder Quality} 
\begin{figure}[!htb]
    \centering  
    \includegraphics[width=\columnwidth, trim={3cm 1cm 3cm 1cm},
        clip]{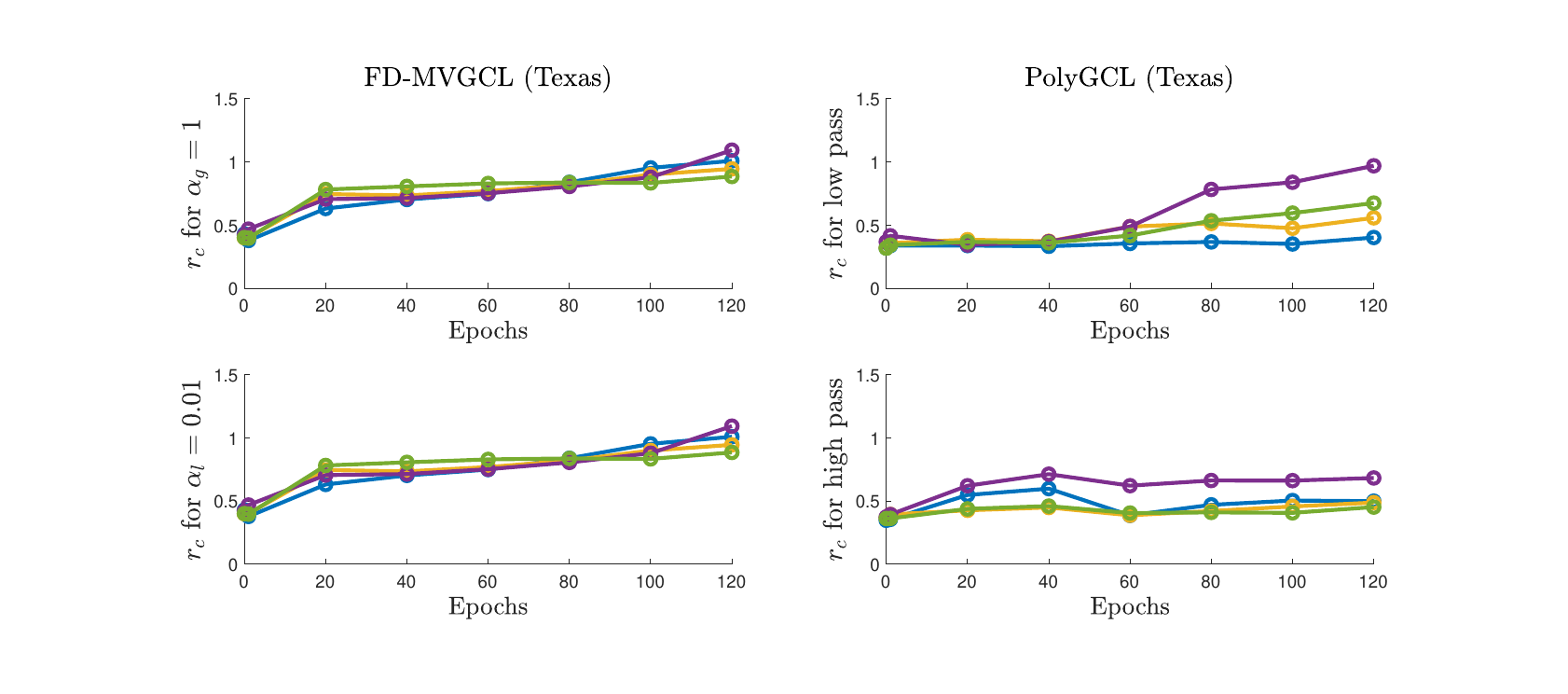}
    \hfill
    \includegraphics[width=\columnwidth, trim={3cm 1cm 3cm 1cm},
        clip]{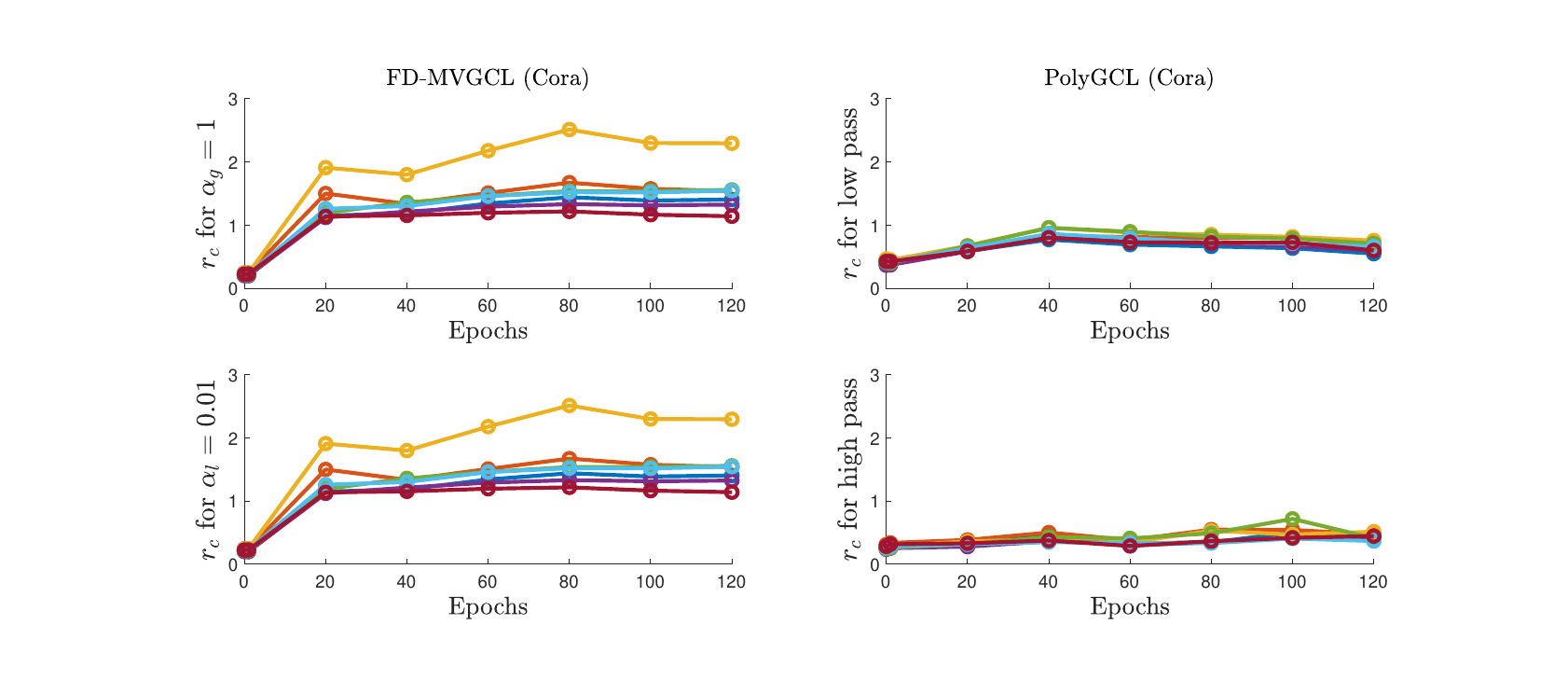}
    \caption{The variation in the ratio $r_c$ during training. Each line plot corresponds to a label class. The ratio $r_c$ for the input features is shown at epoch $0$. We see that the ratio generally increases at the beginning of the training and stabilizes. For example, FD-MVGCL produces higher $r_c$ values than the strong benchmark PolyGCL \cite{chen2024polygcl}, suggesting better clustered embeddings.} 
    \label{fig:wsc}
\end{figure}

The performance of the model improves when the encoders effectively cluster similar nodes. Our approach leverages the proven capabilities of FDE-based encoders in supervised settings \cite{chamberlain2021grand,KanZhaDin:C24frond}. To evaluate their unsupervised clustering ability, we compute the following metrics for each label class $c$: the average intra-class feature distance $d_c^{\mathrm{intra}}$ among nodes within the same class, and the average inter-class feature distance $d_c^{\mathrm{inter}}$ between nodes of class $c$ and nodes of other classes. The clustering quality is quantified by the ratio $r_c = d_c^{\mathrm{inter}}/d_c^{\mathrm{intra}}$, where higher values indicate better clustering. From \cref{fig:wsc}, we observe that our proposed FDE-based encoder yields consistently higher $r_{c}$ values than PolyGCL, whose two filter-based encoders (low-pass and high-pass, analogous to $\alpha_{g}=1$ and $\alpha_{l}=0.01$, respectively) are used for comparison. These results demonstrate that FDE-based encoders produce more discriminative and better-clustered embeddings across both homophilic and heterophilic graphs. 

\subsection{The adaptive FD-MVGCL model}
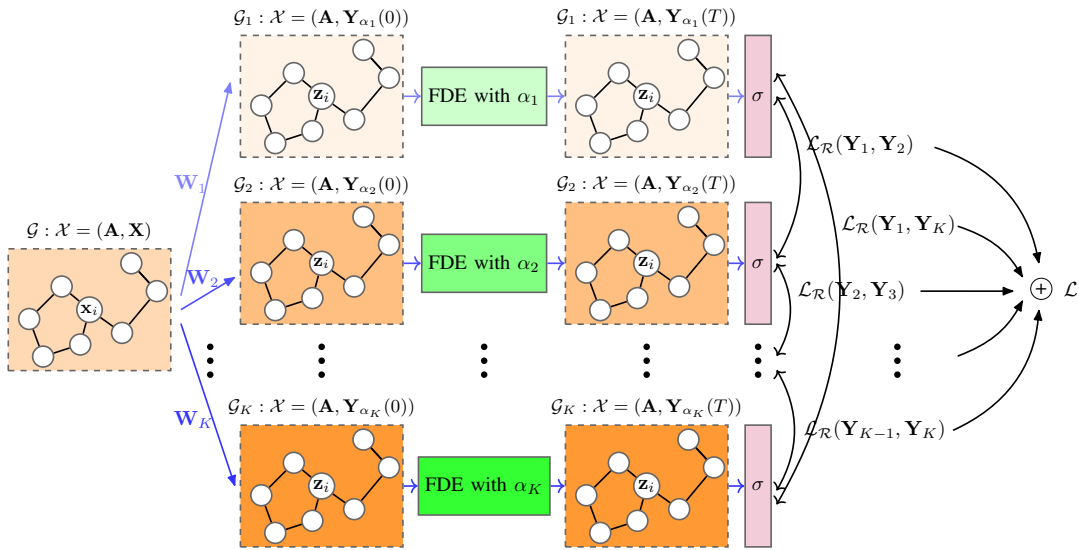
\begin{figure*}[!th]   
\centering
\resizebox{0.8\textwidth}{!}{
    \begin{tikzpicture}[
        scale = 0.8,
        node distance=1.8cm and 2.4cm,
        roundnode/.style={circle, draw=black!60, fill=white!5, thick, minimum size=3.6mm, inner sep=0.3mm, text centered},
        localcircle/.style = {draw = blue!50, dashed, thick, ellipse},
        midcircle/.style = {draw = blue!70, dashed, thick, ellipse},
        globalcircle/.style = {draw = blue!80, dashed, thick, ellipse},
        lightrect/.style={rectangle, dashed, draw=black!60, fill=orange!10, thick, text centered, minimum height=2.5cm, minimum width=3cm},
        originrect/.style={rectangle, dashed, draw=black!60, fill=orange!30, thick, text centered, minimum height=2.5cm, minimum width=3cm},
        midrect/.style={rectangle, dashed, draw=black!60, fill=orange!50, thick, text centered, minimum height=2.5cm, minimum width=3cm},
        darkrect/.style={rectangle, dashed, draw=black!60, fill=orange!80, thick, text centered, minimum height=2.5cm, minimum width=3cm},
        localmodelrect/.style ={rectangle, draw=black!60, fill=green!20, thick, text centered, minimum height=2.5cm, minimum width=3cm},
        midmodelrect/.style ={rectangle, draw=black!60, fill=green!50, thick, text centered, minimum height=2.5cm, minimum width=3cm},
        globalmodelrect/.style ={rectangle, draw=black!60, fill=green!80, thick, text centered, minimum height=2.5cm, minimum width=3cm},
        actrect/.style ={rectangle, draw=black!60, fill=purple!20, thick, text centered, minimum height=2.5cm, minimum width=3cm},
        arrow/.style={-{Latex[round]}, thick},
        dashedarrow/.style={-{Latex[round]}, thick, dashed},
        doublearrow/.style={<->, thick}
    ]

    \node[rectangle, originrect, minimum width=2.8cm, minimum height=2.1cm] (rect) at (0,1) {};

    \node[roundnode] (xi) at (0, 1) {\small{$\bx_i$}};

    \node[roundnode] (n1) at (-1.1+0.5, 0.8-0.3+1) {};
    \node[roundnode] (n2) at (1.5-0.6, 1+1) {};
    \node[roundnode] (n3) at (-1.8+0.5, -0.5+0.3+1) {};
    \node[roundnode] (n4) at (1.8-0.35, 1-0.6+1) {};
    \node[roundnode] (n5) at (1.7-1, -1+0.5+1) {};
    \node[roundnode] (n6) at (-0.5+0.3, -1.2+0.45+1) {};
    \node[roundnode] (n7) at (-1, -1.01+1) {};

    \draw[thick] (xi) -- (n1);
    \draw[thick] (xi) -- (n5);
    \draw[thick] (xi) -- (n6);

    \draw[thick] (n1) -- (n3);
    \draw[thick] (n2) -- (n4);
    \draw[thick] (n4) -- (n5);
    \draw[thick] (n2) -- (n4);
    \draw[thick] (n6) -- (n7);
    \draw[thick] (n3) -- (n7);
    \node[above=0.001cm of rect.north] {\small $\calG:\calX = (\bA, \bX)$};


    \node[rectangle, lightrect, minimum width=2.8cm, minimum height=2.1cm] (rectl) at (5,3.8+1.8) {};

    \node[roundnode] (xi) at (5, 3.8+1.8) {$\bz_i$};

    \node[roundnode] (n1) at (-1.1+5.5, 0.8-0.3+3.8+1.8) {};
    \node[roundnode] (n2) at (1.5-0.6+5, 1+3.8+1.8) {};
    \node[roundnode] (n3) at (-1.8+5.5, -0.5+0.3+3.8+1.8) {};
    \node[roundnode] (n4) at (6.8-0.35, 1-0.6+3.8+1.8) {};
    \node[roundnode] (n5) at (1.7-1+5, -1+0.5+3.8+1.8) {};
    \node[roundnode] (n6) at (-0.5+0.3+5, -1.2+0.45+3.8+1.8) {};
    \node[roundnode] (n7) at (-1+5, -1.01+3.8+1.8) {};

    \draw[thick] (xi) -- (n1);
    \draw[thick] (xi) -- (n5);
    \draw[thick] (xi) -- (n6);

    \draw[thick] (n1) -- (n3);
    \draw[thick] (n2) -- (n4);
    \draw[thick] (n4) -- (n5);
    \draw[thick] (n2) -- (n4);
    \draw[thick] (n6) -- (n7);
    \draw[thick] (n3) -- (n7);

    \node[above=0.001cm of rectl.north] {\small $\calG_{1}:\calX = (\bA,\bY_{\alpha_1}(0))$};


    \node[rectangle, midrect, minimum width=2.8cm, minimum height=2.1cm] (rectm) at (5,2) {};

    \node[roundnode] (xi) at (5, 2) {\small{$\bz_i$}};

    \node[roundnode] (n1) at (-1.1+0.5+5, 0.8-0.3+2) {};
    \node[roundnode] (n2) at (1.5-0.6+5, 1+2) {};
    \node[roundnode] (n3) at (-1.8+0.5+5, -0.5+0.3+2) {};
    \node[roundnode] (n4) at (6.8-0.35, 1-0.6+2) {};
    \node[roundnode] (n5) at (1.7-1+5, -1+0.5+2) {};
    \node[roundnode] (n6) at (-0.5+5.3, -1.2+0.45+2) {};
    \node[roundnode] (n7) at (-1+5, -1.01+2) {};

    \draw[thick] (xi) -- (n1);
    \draw[thick] (xi) -- (n5);
    \draw[thick] (xi) -- (n6);

    \draw[thick] (n1) -- (n3);
    \draw[thick] (n2) -- (n4);
    \draw[thick] (n4) -- (n5);
    \draw[thick] (n2) -- (n4);
    \draw[thick] (n6) -- (n7);
    \draw[thick] (n3) -- (n7);
    \node[above=0.001cm of rectm.north] {\small $\calG_{2}:\calX = (\bA,\bY_{\alpha_2}(0))$};

    \fill (2.6,-0.1) circle (2pt);
    \fill (2.6,0.2) circle (2pt);
    \fill (2.6,-0.4) circle (2pt);
    
    \fill (5,-0.1) circle (2pt);
    \fill (5,0.2) circle (2pt);
    \fill (5,-0.4) circle (2pt);

    \fill (8.5,-0.1) circle (2pt);
    \fill (8.5,0.2) circle (2pt);
    \fill (8.5,-0.4) circle (2pt);

    \fill (12,-0.1) circle (2pt);
    \fill (12,0.2) circle (2pt);
    \fill (12,-0.4) circle (2pt);

    \fill (14.4,-0.1) circle (2pt);
    \fill (14.4,0.2) circle (2pt);
    \fill (14.4,-0.4) circle (2pt);

    \fill (17.4,-0.1) circle (2pt);
    \fill (17.4,0.2) circle (2pt);
    \fill (17.4,-0.4) circle (2pt);


    \node[rectangle, darkrect, minimum width=2.8cm, minimum height=2.1cm] (rectg) at (5,-2.8) {};

    \node[roundnode] (xi) at (5, -2.8) {$\bz_i$};

    \node[roundnode] (n1) at (-1.1+0.5+5, 0.8-0.3-2.8) {};
    \node[roundnode] (n2) at (1.5-0.6+5, 1-2.8) {};
    \node[roundnode] (n3) at (-1.8+0.5+5, -0.5+0.3-2.8) {};
    \node[roundnode] (n4) at (6.8-0.35, 1-0.6-2.8) {};
    \node[roundnode] (n5) at (1.7-1+5, -1+0.5-2.8) {};
    \node[roundnode] (n6) at (-0.5+5.3, -1.2+0.45-2.8) {};
    \node[roundnode] (n7) at (-1+5, -1.01-2.8) {};

    \draw[thick] (xi) -- (n1);
    \draw[thick] (xi) -- (n5);
    \draw[thick] (xi) -- (n6);

    \draw[thick] (n1) -- (n3);
    \draw[thick] (n2) -- (n4);
    \draw[thick] (n4) -- (n5);
    \draw[thick] (n2) -- (n4);
    \draw[thick] (n6) -- (n7);
    \draw[thick] (n3) -- (n7);
    \node[above=0.01cm of rectg.north] {\small $\calG_{K}:\calX = (\bA,\bY_{\alpha_K}(0))$};


    \node[localmodelrect, minimum width=2cm, minimum height=1cm] (rect1) at (8.5, 3.8+1.8) {FDE with $\alpha_{1}$};
    \node[midmodelrect, minimum width=2cm, minimum height=1cm] (rect2) at (8.5, 2) {FDE with $\alpha_{2}$};
    \node[globalmodelrect, minimum width=2cm, minimum height=1cm] (rect3) at (8.5, -2.8) {FDE with $\alpha_{K}$};

    \node[rectangle, lightrect, minimum width=2.8cm, minimum height=2.1cm] (rect5) at (12,3.8+1.8) {};
    \node[above=0.01cm of rect5.north] {\small $\calG_{1}:\calX = (\bA,\bY_{\alpha_1}(T))$};
    \node[roundnode] (xi) at (12, 3.8+1.8) {$\bz_i$};

    \node[roundnode] (n1) at (-1.1+5.5+7, 0.8-0.3+3.8+1.8) {};
    \node[roundnode] (n2) at (1.5-0.6+5+7, 1+3.8+1.8) {};
    \node[roundnode] (n3) at (-1.8+5.5+7, -0.5+0.3+3.8+1.8) {};
    \node[roundnode] (n4) at (6.8-0.35+7, 1-0.6+3.8+1.8) {};
    \node[roundnode] (n5) at (1.7-1+5+7, -1+0.5+3.8+1.8) {};
    \node[roundnode] (n6) at (-0.5+0.3+5+7, -1.2+0.45+3.8+1.8) {};
    \node[roundnode] (n7) at (-1+5+7, -1.01+3.8+1.8) {};

    \draw[thick] (xi) -- (n1);
    \draw[thick] (xi) -- (n5);
    \draw[thick] (xi) -- (n6);

    \draw[thick] (n1) -- (n3);
    \draw[thick] (n2) -- (n4);
    \draw[thick] (n4) -- (n5);
    \draw[thick] (n2) -- (n4);
    \draw[thick] (n6) -- (n7);
    \draw[thick] (n3) -- (n7);


    \node[rectangle, midrect, minimum width=2.8cm, minimum height=2.1cm] (rect6) at (12,2) {};
    \node[above=0.01cm of rect6.north] {\small $\calG_{2}:\calX = (\bA,\bY_{\alpha_2}(T))$};
    \node[roundnode] (xi) at (12, 2) {$\bz_i$};

    \node[roundnode] (n1) at (-1.1+5.5+7, 0.8-0.3+2) {};
    \node[roundnode] (n2) at (1.5-0.6+5+7, 1+2) {};
    \node[roundnode] (n3) at (-1.8+5.5+7, -0.5+0.3+2) {};
    \node[roundnode] (n4) at (6.8-0.35+7, 1-0.6+2) {};
    \node[roundnode] (n5) at (1.7-1+5+7, -1+0.5+2) {};
    \node[roundnode] (n6) at (-0.5+0.3+5+7, -1.2+0.45+2) {};
    \node[roundnode] (n7) at (-1+5+7, -1.01+2) {};

    \draw[thick] (xi) -- (n1);
    \draw[thick] (xi) -- (n5);
    \draw[thick] (xi) -- (n6);

    \draw[thick] (n1) -- (n3);
    \draw[thick] (n2) -- (n4);
    \draw[thick] (n4) -- (n5);
    \draw[thick] (n2) -- (n4);
    \draw[thick] (n6) -- (n7);
    \draw[thick] (n3) -- (n7);


    \node[rectangle, darkrect, minimum width=2.8cm, minimum height=2.1cm] (rect7) at (12,-2.8) {};
    \node[above=0.01cm of rect7.north] {\small $\calG_{K}:\calX = (\bA,\bY_{\alpha_K}(T))$};
    \node[roundnode] (xi) at (12, -2.8) {$\bz_i$};

    \node[roundnode] (n1) at (-1.1+5.5+7, 0.8-0.3-2.8) {};
    \node[roundnode] (n2) at (1.5-0.6+5+7, 1-2.8) {};
    \node[roundnode] (n3) at (-1.8+5.5+7, -0.5+0.3-2.8) {};
    \node[roundnode] (n4) at (6.8-0.35+7, 1-0.6-2.8) {};
    \node[roundnode] (n5) at (1.7-1+5+7, -1+0.5-2.8) {};
    \node[roundnode] (n6) at (-0.5+0.3+5+7, -1.2+0.45-2.8) {};
    \node[roundnode] (n7) at (-1+5+7, -1.01-2.8) {};

    \draw[thick] (xi) -- (n1);
    \draw[thick] (xi) -- (n5);
    \draw[thick] (xi) -- (n6);

    \draw[thick] (n1) -- (n3);
    \draw[thick] (n2) -- (n4);
    \draw[thick] (n4) -- (n5);
    \draw[thick] (n2) -- (n4);
    \draw[thick] (n6) -- (n7);
    \draw[thick] (n3) -- (n7);


    \node[rectangle, actrect, minimum width=0.4cm, minimum height=2.1cm] (rect8) at (14.4, 3.8+1.8) {$\sigma$};
    \node[rectangle, actrect, minimum width=0.4cm, minimum height=2.1cm] (rect9) at (14.4, 2) {$\sigma$};
    \node[rectangle, actrect, minimum width=0.4cm, minimum height=2.1cm] (rect10) at (14.4, -2.8) {$\sigma$};








    \draw[->, thick, draw=blue!50] ($(rect1.east)+(0,0)$) to ($(rect5.west)+(0,0)$);
    \draw[->, thick, draw=blue!70] ($(rect2.east)+(0,0)$) to ($(rect6.west)+(0,0)$);
    \draw[->, thick, draw=blue!80] ($(rect3.east)+(0,0)$) to ($(rect7.west)+(0,0)$);

    \draw[->, thick, draw=blue!50] ($(rectl.east)+(0,0)$) to ($(rect1.west)+(0,0)$);
    \draw[->, thick, draw=blue!70] ($(rectm.east)+(0,0)$) to ($(rect2.west)+(0,0)$);
    \draw[->, thick, draw=blue!80] ($(rectg.east)+(0,0)$) to ($(rect3.west)+(0,0)$);

    \draw[->, thick, draw=blue!50] ($(rect5.east)+(0,0)$) to ($(rect8.west)+(0,0)$);
    \draw[->, thick, draw=blue!70] ($(rect6.east)+(0,0)$) to ($(rect9.west)+(0,0)$);
    \draw[->, thick, draw=blue!80] ($(rect7.east)+(0,0)$) to ($(rect10.west)+(0,0)$);

    \draw[arrow, draw=blue!50] ($(rect.east)+(0.2,0.3)$) -- ($(rectl.west)-(0.2,-0.3)$) node[pos=0.2, above=26pt] {\textcolor{blue!50}{$\bW_{1}$}};
    \draw[arrow, draw=blue!70] ($(rect.east)+(0.2,0)$) -- ($(rectm.west)-(0.1,0.3)$) node[pos=0.4, above=1pt] {\textcolor{blue!70}{$\bW_{2}$}};
    \draw[arrow, draw=blue!80] ($(rect.east)+(0.2,-0.3)$) -- ($(rectg.west)-(0.1,0)$) node[pos=0.24, below=20pt] {\textcolor{blue!80}{$\bW_{K}$}};

    \draw[doublearrow] ($(rect8.east)+(0.1,0)$) to[bend left=30] node[pos=0.3, right=1pt] {$\calL_{\calR}(\bY_1,\bY_2)$}($(rect9.east)+(0.1,0.1)$);
    \draw[doublearrow] ($(rect9.east)+(0.1,0)$) to[bend left=30] node[pos=0.3, right=1pt] {$\calL_{\calR}(\bY_2,\bY_3)$}($(14.7,0)+(0.1,0)$);
    \draw[doublearrow] ($(14.7,-0.2)+(0.1,-0.1)$) to[bend left=30] node[pos=0.5, right=1pt] {$\calL_{\calR}(\bY_{K-1},\bY_K)$}($(rect10.east)+(0.1,-0.1)$);
    \draw[doublearrow] ($(rect8.east)+(0.1,0.5)$) to[bend left=30] node[pos=0.35, right=-0.1pt] {$\calL_{\calR}(\bY_1,\bY_K)$}($(rect10.east)+(0.1,-0.4)$);

    \node (plus) [draw, circle, inner sep=1pt] at (20.5,1.45) {+};

    \node[right=2pt of plus] {$\calL$};

    \draw[arrow] ($(rect8.east)+(3.5,-1.1)$) to[bend left=30] ($(20.45,2)$); 
    \draw[arrow] ($(rect9.east)+(3.2,-0.6)$) to ($(19.9,1.4)$); 
    \draw[arrow] ($(rect10.east)+(3.9,1.2)$) to[bend right=30] ($(20.45,1)$); 
    \draw[arrow] ($(rect10.east)+(4,5.6)$) to[bend left=20] ($(20.1,1.7)$); 
    \draw[arrow] ($(rect10.east)+(4,2.8)$) to[bend right=20] ($(20.1,1.2)$); 

    \end{tikzpicture}
    }
    \caption{Overview of the proposed adaptive FD-MVGCL framework. The fractional orders are selected such that $0<\alpha_{1}<\alpha_{2}<\dots<\alpha_{K}$, enabling contrastive learning across multiple distinct views.}
    \label{fig:framework}
\end{figure*}

We now present the full details of the proposed adaptive FD-MVGCL model. Given a graph $\calG$ with adjacency matrix $\bA$ and initial node features $\bX$, we denote the input as $\calX = (\bA, \bX)$. 
Our objective is to learn multiple FDE encoders $f_{\theta_{k}}$ for $k \geq 2$ in an unsupervised manner, with each encoder capturing structural patterns at a specific scale. 
In the adaptive FD-MVGCL model, all encoders $f_{\theta_{k}}$ share the same architecture (see \cref{fig:framework}). For clarity, we describe the structure of a single encoder below:
\begin{itemize}
    \item \textbf{Feature projection:} Apply a linear transformation with learnable weights $\bW_{k}$ to the input features, i.e., $\bZ_{k} = \bW_{k}\bX$. This step typically increases the feature dimension to enhance the model's expressiveness and ability to capture complex patterns.
    \item \textbf{Fractional diffusion:} Initialize the diffusion process with $\bY_{\alpha_{k}}(0) = \bZ_{k}$, where $\alpha_{k}$ is the fractional order. Compute $\bY_{\alpha_{k}}(t)$ using the FDE-based encoder, as described in \cref{eq.frond_main}.
    \item \textbf{Activation and output:} Stop the diffusion process at time $T$ and apply a non-linear activation function $\sigma$ (e.g., ReLU) to obtain the final output, denoted by $\bY_{k} = \sigma(\bY_{\alpha_{k}}(T))$.
\end{itemize}
The model parameters $\theta_k$ include the learnable weight matrix $\bW_k$, the fractional order $\alpha_{k}$, and the diffusion time $T$. Unlike prior FDE-based models such as FROND \cite{KanZhaDin:C24frond} and DRAGON \cite{ZhaKanJi:C24}, which rely on manually tuning $\alpha$ via grid search, we treat the $\alpha$ values as \textbf{learnable parameters}. This allows the model to adaptively generate informative views tailored to the input data.

We utilize $K$ encoders $f_{\theta_{k}}$, $k=1,2,\dots,K$, each generating an output $\bY_{k}$. To optimize the model parameters, we employ a \emph{contrastive loss} $\calL_0(\bY_k, \bY_{k'})$, where $k'= k+1 \bmod K$, i.e., this loss is applied to each pair of consecutive outputs $(\bY_{k},\bY_{k+1})$, with wrap-around from $\bY_{K}$ to $\bY_{1}$.
For simplicity, $\calL_0$ is defined as the mean cosine similarity, referred to as \emph{cosmean} (see \cref{cosmean_loss}). This consecutive-pair scheme balances representational alignment and computational efficiency by enforcing consistency only between successive views. Additionally, the loss applied to $(\bY_{K},\bY_1)$ ensures global consistency across all encoders. This approach reduces the computational complexity from comparing all possible pairs ($\calO(K^2)$) to only consecutive pairs ($\calO(K)$), maintaining scalability without compromising representation quality.

\subsubsection{Model/View Collapse} As discussed in \cref{sec:ce}, directly minimizing $\calL_0$ can lead to view collapse, where two encoders produce nearly identical representations. To address this, we leverage the observation in \cref{fig:wsp} that both $\bY_k$ and $\bY_{k'}$ exhibit dominant components. Let $\bc_k$ and $\bc_{k'}$ represent the unit directional vectors of these components. To prevent collapse, we introduce a penalty term that discourages alignment between these vectors by incorporating their inner product $\langle \bc_k, \bc_{k'} \rangle$. The resulting \emph{regularized cosmean} loss is defined as:
\begin{align*}
\calL_{\calR}(\bY_k, \bY_{k'}) = \calL_0(\bY_k, \bY_{k'}) + \eta \abs*{\langle \bc_k, \bc_{k'} \rangle},
\end{align*}
where $\eta$ is a regularization weight. This regularization term encourages diversity between the two representations, ensuring they remain distinct without relying on negative samples, unlike traditional contrastive objectives \cite{Teng2023GraphACL,chen2024polygcl}. The impact of this regularization is further analyzed in \cref{subsec.loss}. Finally, the overall contrastive loss across all encoders is given by:
\begin{align}
\label{eq.loss}
\calL = \sum_{k=1}^{K} \calL_{\calR}\parens*{\bY_k, \bY_{k'}}.
\end{align}

\subsubsection{Encoder reduction strategy}
To manage training costs as $K$ increases, we propose an encoder reduction strategy (\cref{alg:AVLA}). Initially, $K$ encoders are created with fractional orders $\{\alpha_k\}_{k=1}^K$ uniformly sampled from the range $(0.01, 1]$. Both the weights $\bW_k$ and the fractional orders $\alpha_k$ are optimized jointly using gradient descent on the loss $\calL$ defined in \cref{eq.loss}. After each training phase, if any $\alpha_k$ values are too similar in log-scale, only one is randomly retained, while the corresponding encoders are reinitialized. The number of encoders $K$ is then reduced accordingly. This process is repeated until all $\alpha_k$ values are sufficiently distinct. The final number of encoders is denoted by $\tilde{K}$. 

Finally, for downstream tasks, we compute a weighted average of the multiple views as $\bY = \sum_{k=1}^{\tilde{K}}\beta_{k}\bY_{k}$, where $\sum_{k=1}^{\tilde{K}}\beta_{k}=1$. The weights $\beta_{k}$ are tuned on a validation set for each downstream task. 

\begin{algorithm}[!tb]
\caption{Adaptive View Learning Algorithm (AVLA)}
\label{alg:AVLA}
\textbf{Input}: Initial number of encoders as $K$, initial fractional orders $\{\alpha_{k}^{(0)}\}_{k=1}^{K}$, initial weights $\{\bW_k^{(0)}\}_{k=1}^{K}$ randomly, diffusion step $T$, learning rate $\eta$ and $\eta_{\alpha_{k}}$, number of training epochs $N$, threshold $\epsilon$ and $\delta$. \\
\textbf{Output}: Optimized number of encoders $\tilde{K}$, fractional orders $\{\tilde{\alpha}_{k}\}_{k=1}^{\tilde{K}}$
\begin{algorithmic}[1] 
\STATE Set \textbf{Repeat} $\leftarrow$ \texttt{True}
\WHILE{\textbf{Repeat} is \texttt{True}}
\FOR{$e=1$ to $N$}
\STATE Compute $\bY_{k}(T)$ via \cref{eq.frond_main} for each encoder $k$
\STATE Compute contrastive loss $\calL$ via \cref{eq.loss}
\STATE Update $\bW_{k}^{(e)} \leftarrow \bW_{k}^{(e-1)}-\eta \nabla_{\bW_{k}} \calL$
\STATE Update $\alpha_{k}^{(e)} \leftarrow \alpha_{k}^{(e-1)}-\eta_{\alpha_{k}} \nabla_{\alpha_{k}} \calL$
\IF {$\alpha_k^{(e)} \notin (0, 1]$}
\STATE $\alpha_k^{(e)} \leftarrow \min(1, \max(\alpha_k^{(e)}, \epsilon))$
\ENDIF
\ENDFOR
\STATE Let $\bar{\balpha} = \big\{ \alpha_i \in \{\alpha_k^{(N)}\}_{k=1}^K \;\big| $
\STATE \hspace{1.5cm} $\max_{\alpha_j, \alpha_\ell \in \bar{\balpha}} \left| \log(\alpha_j) - \log(\alpha_\ell) \right| < \delta \big\}$
\STATE Let $\tilde{\balpha} = \left( \{\alpha_k^{(N)}\}_{k=1}^{K} \setminus \bar{\balpha} \right) \cup \{ \alpha_r \}$, $\alpha_r \sim \text{Unif}(\bar{\balpha})$
\IF {$|\tilde{\balpha}|< K$}
\STATE $K \leftarrow |\tilde{\balpha}|$
\STATE Reinitialize $\{\alpha_k^{(0)}\}_{k=1}^{K} \leftarrow \tilde{\balpha}$ and $\{\bW_k^{(0)}\}_{k=1}^{K}$
\STATE \textbf{Repeat} $\leftarrow$ \texttt{True}
\ELSE 
\STATE $\tilde{K} = K$ and $\{\tilde{\alpha}_{k}\}_{k=1}^{K} = \tilde{\balpha}$
\STATE \textbf{Repeat} $\leftarrow$ \texttt{False}
\ENDIF
\ENDWHILE
\STATE \textbf{return} $\tilde{K}$ and $\{\tilde{\alpha}_k\}_{k=1}^{\tilde{K}}$
\end{algorithmic}
\end{algorithm}

\subsection{Stability}
Graph data in real-world applications often contain noisy node features or missing edges \cite{Lee24}. To assess the robustness of FD-MVGCL, we perform a unified stability analysis that accounts for perturbations in input features, model parameters, and graph structure. \cref{theo:unified_perturbation} establishes the stability of the FDE encoder, while the overall stability of the model is formalized in \cref{Theo.stab_fd_gcl}. Detailed proofs are provided in \cref{sec:stability}.

\begin{Theorem}\label{theo:unified_perturbation}
Let $\bY(t)$ denote the solution to the original FDE encoder in \cref{eq.frond_main}, and let $\tilde{\bY}(t)$ be the solution to the perturbed system:
\begin{align}
\begin{aligned}
\label{eq.fde_per}
D_t^\alpha \tilde{\bY}(t) &= \tilde\calF(\tilde{\bW},\tilde{\bY}(t)),  \\
\tilde{\bY}(0) &= \tilde{\bY}_{0}.
\end{aligned}
\end{align}
Define the total perturbation magnitude as
\begin{align}
\label{eq.comb_per}
\varepsilon := \norm{\bY_{0} - \tilde{\bY}_{0}} + \sup_{\bY\in A} \norm{\calF(\bW,{\bY}) - \tilde\calF(\tilde{\bW},{\bY})},
\end{align}
where $A$ denotes a compact set where the solutions of both systems are well defined. Suppose that $\calF$ is Lipschitz continuous with constant $L$. Then, for sufficiently small $\varepsilon$, there exist a sufficiently large positive constant $T$ such that both the functions \(\bY(t)\) and \(\tilde{\bY}(t)\) are defined on \([0, T]\), and the following perturbation bound holds:
\begin{align}
  \norm{\bY(t)-\tilde{\bY}(t)} = O\left(\varepsilon T^{\alpha-1}\right),
\end{align}
for $t\in [0,T]$. 
\end{Theorem}

\Cref{theo:unified_perturbation} extends both Theorem 6.20 and Theorem 6.21 from \cite{diethelm2010frationalde} by providing a unified bound that accounts for both initial state and functional perturbations. When the perturbation is solely due to the initial state, i.e., $\varepsilon := \norm{\bY_{0} - \tilde{\bY}_{0}}$, the theorem simplifies to Theorem 6.20. Similarly, when the perturbation arises only from the governing function, i.e., $\varepsilon := \sup_{\bY\in A} \norm{\calF(\bW,{\bY}) - \tilde\calF(\tilde{\bW},{\bY})}$, it reduces to Theorem 6.21. Thus, \cref{theo:unified_perturbation} unifies these scenarios, offering a comprehensive stability result for simultaneous perturbations.

\begin{Theorem}
\label{Theo.stab_fd_gcl}
Suppose $0 < \alpha_1 < \alpha_2 < \dots < \alpha_{\tilde{K}} \leq 1$ and let
\begin{align}
\label{eq.solu_fd_gcl}
\bY(t) = \sum_{k=1}^{\tilde{K}} \beta_{k} \bY_{k}(t),
\end{align}
where each $\bY_{k}(t)$ is the solution to the feature dynamics in \cref{eq.frond_main}, denote the solution at evolution time $t$ of the FD-MVGCL model with $K$ encoders.   
Let $\tilde{\bY}(t)$ be the perturbed solution of FD-MVGCL where each encoder is a perturbed system in \cref{eq.fde_per}. 
For sufficiently small combined perturbation $\varepsilon$ in \cref{eq.comb_per}, and a sufficiently large time horizon $T>0$ such that both the functions \(\bY(t)\) and \(\tilde{\bY}(t)\) are defined on \([0, T]\), then we have
\begin{align}
\label{eq.stab_fdgcl}
  \norm{\bY(t)-\tilde{\bY}(t)} = O\left(\varepsilon T^{\alpha_{\tilde{K}}-1}\right)
\end{align}
for $t\in [0,T]$.
\end{Theorem}

\Cref{Theo.stab_fd_gcl} highlights the inherent robustness of FD-MVGCL: smaller $\alpha_{k}$ values result in reduced feature discrepancies in \cref{eq.stab_fdgcl}, making the diffusion process more resilient to perturbations when $\alpha_{k} < 1$. This property underscores FD-MVGCL's robustness against adversarial attacks, as further demonstrated in \cref{sec.robust_study}.

\section{Experiments}

\subsection{Experimental setup}

\subsubsection{Datasets and splits} We evaluate FD-MVGCL on both heterophilic and homophilic datasets. The heterophilic benchmarks include Wisconsin, Cornell, Texas, Actor, Crocodile, and the filtered versions of Squirrel and Chameleon, where duplicate nodes are removed to prevent training–test leakage \cite{platonov2023critical}.\footnote{Note that this typically leads to lower performance of baseline methods in our presented results compared to those reported in the baseline papers.} In addition, we consider three larger heterophilic datasets: Roman-empire (Roman), Amazon-ratings (Amazon), and Arxiv-year. The homophilic datasets include Cora, Citeseer, Pubmed, Computer, Photo, and the large-scale Ogbn-Arxiv. We adopt the standard public splits from \cite{Teng2023GraphACL}, with further details provided in \cref{dataset_statistic}.

\subsubsection{Baselines} We compare FD-MVGCL with recent state-of-the-art unsupervised learning methods: DGI \cite{velickovic2019dgi}, GMI \cite{peng2020gmi}, MVGRL \cite{hassani2020mvgrl}, GRACE \cite{zhu2020GRACE}, CCA-SSG \cite{Zhang2021CCA_SSG}, BGRL \cite{thakoor2022BGRL}, AFGRL \cite{Lee2022AFGRL}, HGRL \cite{Chen2022HGRL}, DSSL \cite{Teng2022DSSL}, SP-GCL \cite{Wang2023SPGCL}, GraphACL \cite{Teng2023GraphACL}, GraphECL \cite{xiaographecl2024}, PolyGCL \cite{chen2024polygcl}, LOHA \cite{Loha2025}, EPAGCL \cite{XuEPAGCL2025}, AFECL \cite{LiAFECL2025} and SDMG \cite{zhu2025sdmg}.

\subsubsection{Evaluation protocol} We mainly evaluate representation quality via node classification using the standard linear protocol: a linear classifier is trained on frozen embeddings, and test accuracy is reported. We further assess FD-MVGCL on the graph classification task (see \cref{graphclassification}) to demonstrate its generalization beyond node-level settings. 

\subsubsection{Setup} For FD-MVGCL, we adopt the basic GRAND form of $\calF$ in \cref{eq.frond_main}. Model parameters are randomly initialized and optimized using Adam. Each experiment is run with $10$ random seeds (as in \cite{Teng2023GraphACL}), and we report the mean and standard deviation. All methods are tuned solely on validation accuracy for fairness. For baselines without results on certain datasets or using nonstandard splits \cite{chen2024polygcl,XuEPAGCL2025,Loha2025}, we reproduce results using the authors’ official code. Implementation and hyperparameter details are in \cref{supp.hyperparameter}. 

\begin{table*}[!htb]
\caption{Node classification results(\%) on heterophilic datasets. The best and the second-best result under each dataset are highlighted in \first{} and \second{}, respectively. OOM refers to out of memory on an NVIDIA A5000 GPU (24GB).} 
\centering
\fontsize{7.5pt}{8pt}\selectfont
\setlength{\tabcolsep}{1.5pt}
\begin{tabular}{lccc|cccccccc|c}
\toprule
Method          & DGI              & CCA-SSG           & BGRL             & DSSL            &SP-GCL            & GraphACL        &PolyGCL            & GraphECL           &LOHA               & EPAGCL           &SDMG              & FD-MVGCL   \\
\midrule
Squirrel        & 40.60$\pm$0.35   & \second{41.23$\pm$1.77}    &  40.13$\pm$2.16  &  37.69$\pm$2.07 & 40.11$\pm$2.20     & 35.51$\pm$2.03  & 33.07$\pm$0.94    & 35.89$\pm$2.62     &34.46$\pm$1.69    & 40.28$\pm$1.59   & 36.82$\pm$0.77    &  \first{41.47$\pm$1.54} \\
Chameleon       & 42.57$\pm$0.71   & 39.46$\pm$3.10    & 42.55$\pm$2.35    & 40.95$\pm$3.35  & 44.49$\pm$2.59   & 38.59$\pm$2.81  & 41.79$\pm$2.45    & 36.11$\pm$1.15     &\second{45.45$\pm$1.83}    & 35.43$\pm$1.28  & 41.55$\pm$0.73 &  \first{45.57$\pm$2.00}\\
Crocodile       & 51.25$\pm$0.51   & 56.77$\pm$0.39    & 53.87$\pm$0.65   & 62.98$\pm$0.51  & 61.72$\pm$0.21   & \second{66.17$\pm$0.24}  & 65.95$\pm$0.59    & 52.52$\pm$3.01     &66.09$\pm$0.69    & 66.14$\pm$0.62   & 65.38$\pm$0.37    & \first{68.41$\pm$0.57}\\
Actor           & 28.30$\pm$0.76   & 27.82$\pm$0.60    & 28.80$\pm$0.54   & 28.15$\pm$0.31  & 28.94$\pm$0.69   & 30.03$\pm$0.13  & 34.37$\pm$0.69    & \second{35.80$\pm$0.89}     &33.69$\pm$0.73    & 30.02$\pm$0.91   & 26.74$\pm$0.13    & \first{35.91$\pm$0.92}\\
Wisconsin       & 55.21$\pm$1.02   & 58.46$\pm$0.96    & 51.23$\pm$1.17   & 62.25$\pm$0.55  & 60.12$\pm$0.39   & 69.22$\pm$0.40  & 76.92$\pm$1.48    & \first{79.41$\pm$0.65}      &74.27$\pm$1.99    & 63.73$\pm$3.95   & 52.68$\pm$1.21    & \second{78.04$\pm$0.08}  \\
Cornell         & 45.33$\pm$6.11   & 52.17$\pm$1.04    & 50.33$\pm$2.29   & 53.15$\pm$1.28  & 52.29$\pm$1.21   & 59.33$\pm$1.48  & 44.11$\pm$1.12    & \second{66.24$\pm$1.82}     &55.35$\pm$1.32    & 53.78$\pm$3.72   & 45.59$\pm$0.67    & \first{72.70$\pm$0.57}  \\
Texas           & 58.53$\pm$2.98   & 59.89$\pm$0.78    & 52.77$\pm$1.98   & 62.11$\pm$1.53  & 59.81$\pm$1.33   & 71.08$\pm$0.34  & 72.51$\pm$1.95    & \second{73.76$\pm$2.02}     &67.75$\pm$2.70    & 68.92$\pm$4.05   & 53.60$\pm$2.67    & \first{80.19$\pm$0.81} \\
Roman           & 63.71$\pm$0.63   & 67.35$\pm$0.61    & 68.66$\pm$0.39   & 71.70$\pm$0.54  & 70.88$\pm$0.35   & \first{74.91$\pm$0.28}  & \second{72.97$\pm$0.25}    & 45.05$\pm$1.57     & OOM      & 47.11$\pm$0.87   & 49.20$\pm$0.51    &  72.45$\pm$0.73\\
Amazon          & 42.72$\pm$0.42   & 41.23$\pm$0.25    & 41.17$\pm$0.25   & 42.12$\pm$0.78  & 42.04$\pm$0.68   & OOM             & \first{44.29$\pm$0.43}    & 36.88$\pm$1.25     & 38.45$\pm$0.20   & OOM              & 43.70$\pm$0.36    & \second{43.86$\pm$0.77}   \\
Arxiv-year      & 39.26$\pm$0.72   & 37.38$\pm$0.41    & 43.02$\pm$0.62   & 45.80$\pm$0.57  & 44.11$\pm$0.35   & \second{47.21$\pm$0.39}  & 43.07$\pm$0.23    & OOM                & OOM              & OOM              &  OOM              & \first{47.35$\pm$0.22} \\
\midrule
\midrule
Avg.\ rank.\ & 8.00             & 7.60              &  7.50            & 6.30            &  6.70            &  \second{4.38}          &  5.20    &      5.00          &    5.43          &  6.33           &    8.22        & \first{1.40} \\
\bottomrule
\end{tabular}
\vspace{0.8mm}
\label{tab:noderesults_heter}
\end{table*}

\begin{table*}[!htb]
\caption{Node classification results(\%) on homophilic datasets.} 
\centering
\fontsize{7pt}{7.5pt}\selectfont
\setlength{\tabcolsep}{1.3pt}
 \begin{tabular}{lccccc|ccccccc|c}
\toprule
Method      &  DGI            &GMI              &MVGRL            &GRACE            &BGRL              &AFGRL            & GraphACL                 &PolyGCL           &LOHA             &EPAGCL          &AFECL     & SDMG            & FD-MVGCL       \\
\midrule
Cora        & 82.30$\pm$0.60  &82.70$\pm$0.20   &82.90$\pm$0.71   &76.80$\pm$0.90   & 82.70$\pm$0.60   &82.31$\pm$0.42   & \second{84.20$\pm$0.31}   & 82.74$\pm$0.14   &81.22$\pm$0.17           & 82.14$\pm$0.89  &82.10$\pm$1.30   &83.60$\pm$0.60     &\first{84.40$\pm$0.04}      \\
Citeseer    & 71.80$\pm$0.70  &73.01$\pm$0.30   &72.61$\pm$0.70   &71.72$\pm$0.62   & 71.10$\pm$0.80   &68.70$\pm$0.30   & \first{73.63$\pm$0.22}   & 71.82$\pm$0.45   &71.89$\pm$0.63   &71.94$\pm$0.57  &67.14$\pm$1.87   &73.20$\pm$0.50     &\second{73.32$\pm$0.41}    \\
Pubmed      & 76.80$\pm$0.60  &80.11$\pm$0.22   &79.41$\pm$0.31   &79.51$\pm$1.10   & 79.60$\pm$0.50   &79.71$\pm$0.21   & \first{82.02$\pm$0.15}   & 77.31$\pm$0.27   &78.09$\pm$0.29   & 81.28$\pm$0.62  &81.20$\pm$1.70   &80.00$\pm$0.40    &\second{81.32$\pm$0.22}     \\ 
Computer    & 83.95$\pm$0.47  &82.21$\pm$0.34   &87.52$\pm$0.11   &86.51$\pm$0.32   & 89.69$\pm$0.37  &89.90$\pm$0.31  & 89.80$\pm$0.25       & 86.54$\pm$0.45   &79.05$\pm$0.32   & 76.81$\pm$0.79  &- -   &\first{90.40$\pm$0.20}             &\second{90.38$\pm$0.86}      \\
Photo       & 91.61$\pm$0.22  &90.72$\pm$0.21   &91.72$\pm$0.10   &92.50$\pm$0.22   & 92.90$\pm$0.30   &93.25$\pm$0.33  & 93.31$\pm$0.19  & 91.45$\pm$0.35   &86.46$\pm$0.41   & 93.05$\pm$0.23  & 89.20$\pm$1.20   &\second{94.10$\pm$0.20}     &\first{94.27$\pm$0.85}      \\
Ogbn-arxiv  & 70.32$\pm$0.25  & OOM             & OOM             &65.10$\pm$0.50   &\second{71.64$\pm$0.24} & OOM      & \first{71.72$\pm$0.26}    & OOM              & OOM            & 69.29$\pm$0.01  &- -   &70.60$\pm$0.20                  & 70.46$\pm$0.06      \\
\midrule 
\midrule
Avg.\ rank.\ & 9.00          & 6.40            &  6.83           &  8.83           &  5.83            &   6.00        &  \first{2.00}    &  8.67              &  9.40           &  6.33          &  9.75          & \second{3.50}  & \first{2.00}\\
\bottomrule
\end{tabular}
\label{tab:noderesults_homo}
\end{table*}

\subsection{Overall performance}
Node classification results on heterophilic and homophilic datasets are reported in \cref{tab:noderesults_heter,tab:noderesults_homo}, respectively. FD-MVGCL consistently achieves state-of-the-art or highly competitive performance across both graph types. On heterophilic datasets, it substantially outperforms baseline methods, securing the best or second-best results on most benchmarks. On homophilic datasets, FD-MVGCL remains competitive with leading approaches such as GraphACL and PolyGCL. We also report the average ranking, computed as the arithmetic mean of its valid per-dataset ranks (excluding OOM and missing results), where lower values indicate better overall performance. FD-MVGCL attains the lowest average rank on heterophilic datasets and shares the lowest rank with GraphACL on homophilic datasets, demonstrating its strong generalization and consistency across diverse graph types. 

The stronger performance gains on heterophilic graphs reflect the inherent difficulty of unsupervised learning in such settings. Unlike homophilic graphs, where simple clustering methods often suffice, heterophilic graphs require more sophisticated multi-scale representations to capture complex structural patterns. These results demonstrate that FD-MVGCL effectively learns expressive embeddings across diverse graph structures by adaptively balancing local and global diffusion scales. Details on the final number of encoders $\tilde{K}$ and the learned fractional orders $\{\tilde{\alpha}_k\}_{k=1}^{\tilde{K}}$ for each dataset are provided in \cref{supp.hyperparameter}.

\begin{table}[!htb]
\caption{Node classification results (\%) across different datasets and parameter configurations.}
\fontsize{8pt}{8pt}\selectfont
\setlength{\tabcolsep}{2.5pt}
\centering
\begin{tabular}{lcc}
\toprule
\multicolumn{1}{c}{Method}                                                       & Cora                             & Cornell               \\
\midrule
                               (i) Two-view                                     & 83.56$\pm$0.22                   & \second{65.14$\pm$0.54}  \\                               
                               (ii) Three-view                                  & \second{83.67$\pm$0.23}          & 62.16$\pm$0.38          \\
                               (iii) Five-view                                  & 83.61$\pm$0.25                   & 63.51$\pm$0.82           \\
\midrule
                              \multirow{1}{*}{(iv) AVLA}                        & \first{84.40$\pm$0.04}           & \first{72.70$\pm$0.57}  \\
                               $\{\tilde{\alpha}_{k}\}_{k=1}^{\tilde{K}}$       & $\{0.001,0.529,0.736,1\}$        & $\{0.001, 0.018, 0.137, 1\}$      \\  
\bottomrule
\end{tabular}
\label{tab:fd-gcl_alpha}
\end{table}

\begin{table}[!htb]
\caption{Classification accuracy w.r.t diffusion depths ($T$) across different datasets.}
\fontsize{8pt}{9pt}\selectfont
\setlength{\tabcolsep}{2.5pt}
\centering
\begin{tabular}{lcccc}
\toprule
$T$                                &5                     & 15                 & 20                & 25                              \\
\midrule
Cora                               &81.48$\pm$0.12    	  & 83.38$\pm$0.37	   & 84.20$\pm$0.26	   & 84.14$\pm$0.07                    \\
Ogbn-arxiv                         &63.84$\pm$0.12    	  & 66.43$\pm$0.14	   & 66.46$\pm$0.08	   & 69.76$\pm$0.04                    \\
\midrule
Cornell                            &62.97$\pm$0.52    	  & 63.70$\pm$0.71	   & 68.38$\pm$0.54	   & 67.97$\pm$0.82                  \\
Wisconsin                          &71.57$\pm$0.19    	  & 74.12$\pm$0.54	   & 78.04$\pm$0.08	   & 78.06$\pm$0.12                     \\
\bottomrule
\end{tabular}
\vspace{0.8mm}
\label{tab:abla_diffu_depth}
\end{table}

\subsection{Ablation studies}

\subsubsection{Multi-view vs two-view}
We evaluate the performance of FD-MVGCL under various configurations of $\{\alpha_{k}\}_{k=1}^{K}$: (i) a fixed two-view setting with distinct fractional orders ($\alpha{1} = 0.01$, $\alpha_{2} = 1$), (ii) a fixed three-view setting ($\alpha_{1} = 0.01$, $\alpha_{2} = 0.51$, $\alpha_{3} = 1$), (iii) a fixed five-view setting ($\alpha_{1} = 0.01$, $\alpha_{2} = 0.26$, $\alpha_{3} = 0.51$, $\alpha_{4} = 0.75$, $\alpha_{5} = 1$), and (iv) AVLA (\cref{alg:AVLA}) that dynamically selects the number and values of $\alpha_k$ during training. These are compared on the Cora and Cornell datasets using classification accuracy, as shown in \cref{tab:fd-gcl_alpha}.

\subsubsection{Feature dimension} In FD-MVGCL model, we apply a linear encoder that typically increases the feature dimension to enhance the model’s expressiveness and ability to capture complex patterns. As verified in \cref{fig:feature_dim}, increasing the feature dimension consistently enhances performance across both homophilic and heterophilic graphs, with particularly notable gains on heterophilic graphs.

\subsubsection{Diffusion depths}  As observed in \cite{KanZhaDin:C24frond}, increasing the diffusion depth $T$ improves the solution accuracy of \cref{eq.frond_main} and generally enhances view diversity. Empirical results in \cref{tab:abla_diffu_depth} validate that larger $T$ (up to a threshold) yields better performance on both homophilic and heterophilic datasets. The selected $T$ values per dataset are listed in \cref{supp.hyperparameter}.

\subsubsection{Different options for \txp{$\calF$}{F}} We further evaluate FD-MVGCL under alternative formulations of $\calF$ in \cref{eq.frond_main}, including CDE, GREAD  and GraphCON. As reported in \cref{tab:fd-gcl_model}, FD-MVGCL consistently achieves competitive performance across these variants, demonstrating its generality and adaptability to diverse diffusion dynamics.

\begin{table}[!htb]
\caption{Node classification results (\%) across different models in FD-MVGCL.}
\fontsize{7pt}{8pt}\selectfont
\centering
\begin{tabular}{llccc}
\toprule
                         & \multicolumn{1}{c}{Model}            & Cora                     & Cornell             & Roman\\
\midrule
\multirow{4}{*}{FD-MVGCL} & GRAND \cite{chamberlain2021grand}   & 84.40$\pm$0.04           & 72.70$\pm$0.57        & 72.47$\pm$0.70 \\                               
                          & CDE   \cite{ZhaKanSon:C23}          & 75.02$\pm$0.26           & 60.54$\pm$3.92        & OOM \\
                          & GREAD  \cite{choi2023gread}         & 81.02$\pm$0.15           & 66.24$\pm$0.67        & 71.48$\pm$0.36 \\
                          & GraphCON \cite{rusch2022graph}      & 80.64$\pm$0.33           & 67.03$\pm$0.33        & 68.20$\pm$0.65 \\
\bottomrule
\end{tabular}
\label{tab:fd-gcl_model}
\end{table}

\subsubsection{Loss functions} \label{subsec.loss}
\begin{figure}[!htb]
    \centering
    \includegraphics[width=0.75\linewidth]{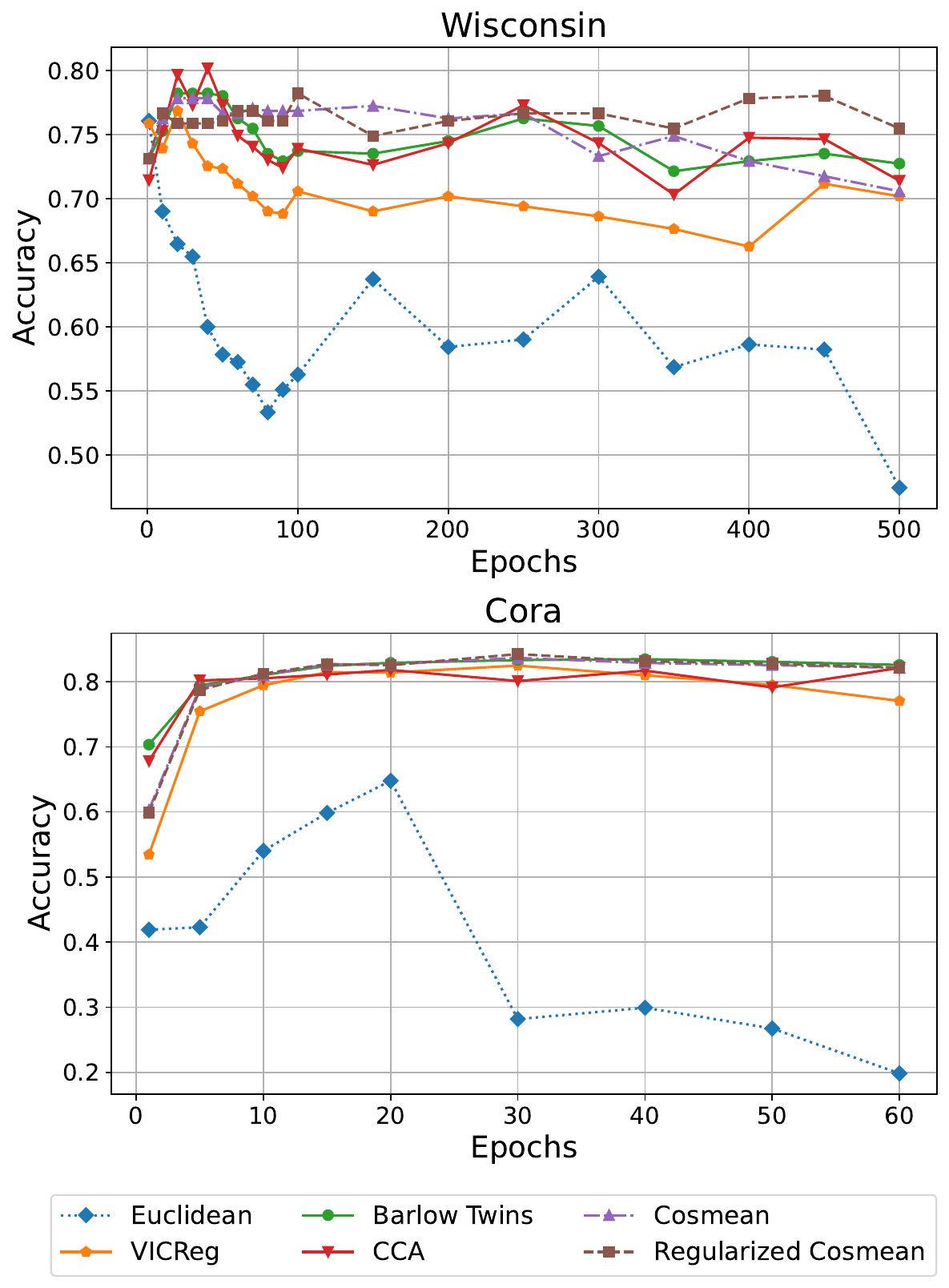}
    \centering
    \caption{Accuracy vs. training epochs for various loss functions on the Wisconsin and Cora datasets.}
    \label{accuracy_loss}
\end{figure}

To assess the performance of FD-MVGCL across various contrastive learning paradigms, we compare several widely adopted \emph{contrastive loss} functions and their impact on two FDE encoder feature representations, where the local view $\bY_{l}$ is obtained with fractional order $\alpha_{l}=0.01$ and and the global view $\bY_{g}$ with $\alpha_{g}=1$. We measure classification accuracy on two benchmark datasets: Cora (homophilic) and Wisconsin (heterophilic). The loss functions considered include Euclidean loss, Cosmean loss, Barlow Twins loss \cite{zbontarbarlowtwin2021}, VICReg loss \cite{bardes2022vicreg}, CCA \cite{Zhang2021CCA_SSG}, and our proposed Regularized Cosmean loss. As illustrated in \cref{accuracy_loss}, the Regularized Cosmean loss consistently outperforms the others, demonstrating stable accuracy throughout the training epochs. This robustness is attributed to its effectiveness in mitigating dimensional collapse, which ensures reliable performance over extended training periods. Detailed definitions of these loss functions and additional results on other datasets can be found in \cref{supp.loss}.

\begin{figure*}[!htb]
\begin{subfigure}{0.328\textwidth}
    \centering
    \includegraphics[width=\linewidth]{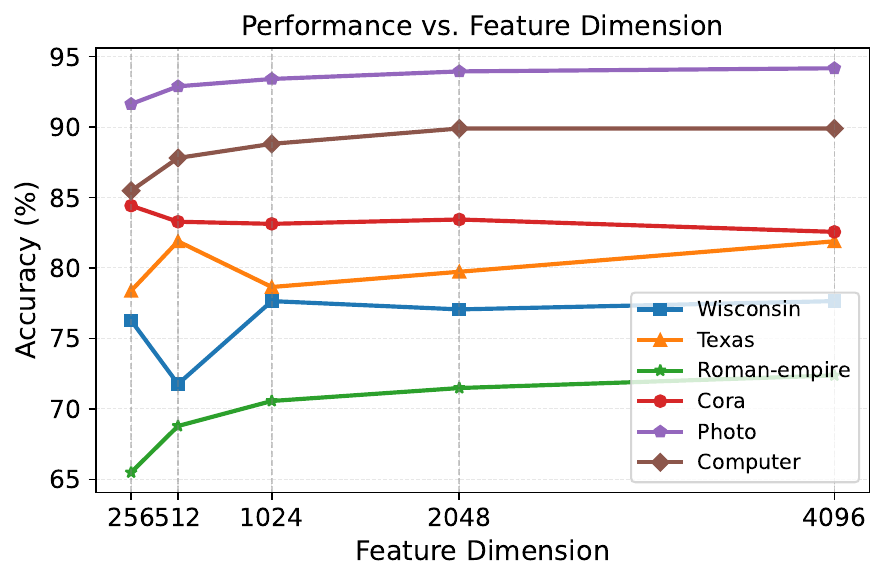}
    \caption{Accuracy vs. feature dimension}
    \label{fig:feature_dim}
\end{subfigure}
\hfill
\begin{subfigure}{0.328\textwidth}
    \centering
    \includegraphics[width=\linewidth]{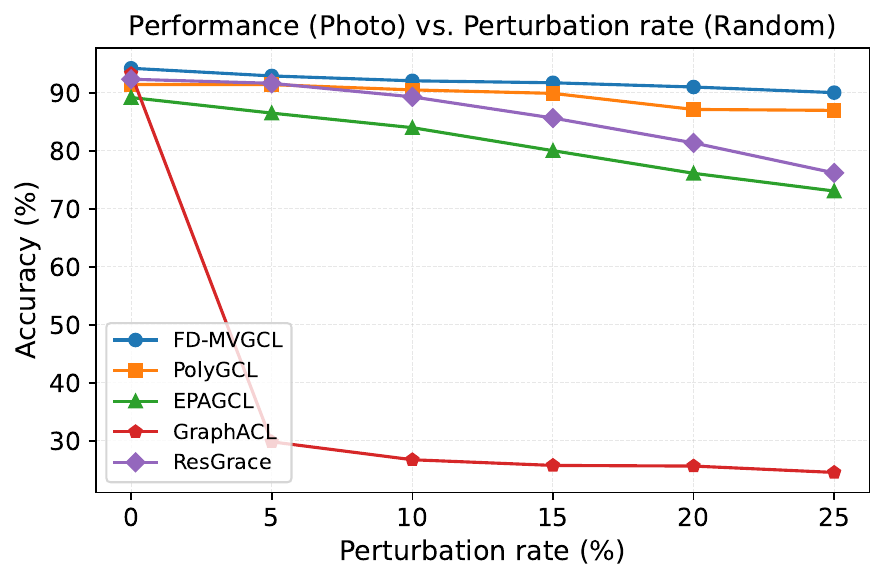}
    \caption{Accuracy vs Random attack rate}
    \label{fig:photo_random_ratio}
\end{subfigure}
\hfill
\begin{subfigure}{0.328\textwidth}
    \centering
    \includegraphics[width=\linewidth]{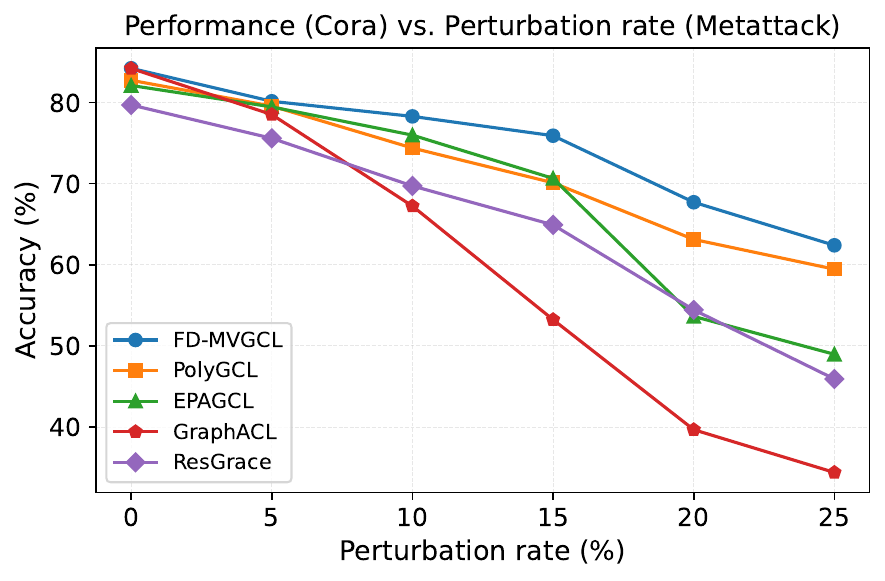}
    \caption{Accuracy vs. Metattack rate}
    \label{fig:cora_meta_ratio}
\end{subfigure}
\centering
\makebox[\textwidth]{\parbox{0.5\textwidth}{%
    \caption{Comparison across feature dimension, Metattack, and Random attack}
    \label{fig:robustness}
}}
\vspace{-3mm}
\end{figure*}

\begin{table*}[!htp]
\caption{Black-box attack robust accuracy (\%) under 20\% perturbation for node classification.}
\label{tab:blackbox_results}
\centering
\fontsize{9pt}{10pt}\selectfont
\setlength{\tabcolsep}{2pt}
\begin{tabular}{c|cc|ccc|cccc|c} 
\toprule
Dataset & Attack                  & FROND             & GCL-Jac       & Ariel        & Res-GRACE      & GraphACL       & PolyGCL        & LOHA        & EPAGCL    & FD-MVGCL    \\ 
\midrule
\multirow{4}{*}{Cora} 
&\emph{clean}                    & 81.25$\pm$0.89     & 76.80$\pm$0.90 & 79.80$\pm$0.60 & 79.70$\pm$1.00 & \second{84.20$\pm$0.31} & 82.74$\pm$0.14 & 81.22$\pm$0.17& 82.10$\pm$1.30  & \first{84.40$\pm$0.04} \\
& Random                         & 78.65$\pm$1.15     & 74.10$\pm$2.00 & 76.30$\pm$0.60 & \second{78.89$\pm$1.17} & 76.92$\pm$0.14 & 78.75$\pm$0.77 & 75.74$\pm$1.14 & 75.33$\pm$0.34  & \first{79.16$\pm$0.12} \\
& PRBCD                          & 79.17$\pm$ 0.57    & 75.30$\pm$1.20 & 75.60$\pm$0.20 & 78.80$\pm$1.16 & \second{80.29$\pm$2.13} & 79.78$\pm$1.08 & 79.70$\pm$0.35 & 77.94$\pm$1.11  & \first{81.06$\pm$1.34} \\
& Metattack                      & 73.42$\pm$0.97     & 67.49$\pm$2.23 & 63.50$\pm$1.50 & 67.82$\pm$1.36 & 53.24$\pm$2.66 & 70.11$\pm$1.53 & \second{74.69$\pm$0.41} & 70.65$\pm$1.27  & \first{80.28$\pm$0.77} \\
& Nettack                        & 80.11$\pm$1.22     & 72.90$\pm$1.39 & 70.30$\pm$0.90 & 76.36$\pm$1.47 & \second{81.39$\pm$0.73} & 81.45$\pm$0.47 & 79.26$\pm$0.86 & 79.54$\pm$0.66  & \first{82.05$\pm$0.55} \\

\midrule

\multirow{4}{*}{Citeseer} 
&\emph{clean}                    & 71.37$\pm$1.34     & 70.52$\pm$0.65 & 50.89$\pm$3.76  & 71.72$\pm$0.62 & \first{73.60$\pm$0.70} & 71.82$\pm$0.45 & 71.95$\pm$0.45 & 71.94$\pm$0.57  & \second{73.32$\pm$0.41} \\
& Random                         & 70.23$\pm$1.40     & 57.26$\pm$4.20 & 44.98$\pm$3.45  & 56.69$\pm$2.63 & 68.13$\pm$0.44 & \second{71.58$\pm$0.24} & \first{71.70$\pm$0.29} & 63.10$\pm$1.08  & 71.23$\pm$0.55 \\
& PRBCD                          & 71.47$\pm$1.29     & 58.30$\pm$4.11 & 46.02$\pm$3.16  & 58.86$\pm$266 & 70.52$\pm$1.16 & 71.19$\pm$0.61 & \first{71.60$\pm$0.63} & 64.54$\pm$2.00  & \second{71.57$\pm$1.17} \\
& Metattack                      & 67.94$\pm$1.42     & 57.51$\pm$5.21  & 36.68$\pm$3.76  & 36.20$\pm$5.62  & 20.50$\pm$0.28 & \first{71.78$\pm$0.42} & 42.99$\pm$4.02 & 47.24$\pm$2.67  & \second{68.99$\pm$2.62} \\
& Nettack                        & 70.05$\pm$1.10      & 59.40$\pm$4.17 & 46.45$\pm$3.16  & 58.18$\pm$2.65 & \first{71.93$\pm$1.10} & 70.33$\pm$0.50 & 71.01$\pm$0.34 & 65.27$\pm$1.21  & \second{71.09$\pm$1.01} \\
\midrule

\multirow{4}{*}{Cornell} 
&\emph{clean}          & \second{63.24$\pm$9.38}    & 42.97$\pm$6.78 & 51.89$\pm$6.71  & 51.08$\pm$5.19 & 59.33$\pm$1.48 & 44.10$\pm$1.12 & 55.35$\pm$1.35 & 53.78$\pm$3.72  & \first{72.70$\pm$0.57}  \\
& Random               & \second{63.24$\pm$7.27}    & 37.30$\pm$4.49 & 40.00$\pm$4.95  & 49.19$\pm$4.15 & 42.97$\pm$8.10 & 43.78$\pm$5.14 & 45.68$\pm$3.51 & 54.32$\pm$6.33  & \first{64.62$\pm$0.41}  \\
& PRBCD                & \second{64.86$\pm$5.27}    & 41.62$\pm$9.83 & 48.38$\pm$6.33  & 48.92$\pm$5.98 & 46.22$\pm$9.66 & 44.59$\pm$4.05 & 51.08$\pm$3.24 & 53.24$\pm$6.74  & \first{67.32$\pm$1.50}  \\
& Metattack            & \second{67.03$\pm$5.51}    & 38.65$\pm$6.63 & 49.73$\pm$7.85  & 49.73$\pm$6.07 & 45.14$\pm$6.87 & 42.43$\pm$4.87 & 48.11$\pm$5.14 & 55.68$\pm$5.86  & \first{67.95$\pm$0.53}  \\
& Nettack              & \second{66.49$\pm$6.53}    & 41.08$\pm$7.03 & 50.54$\pm$6.95  & 49.19$\pm$5.24 & 49.73$\pm$7.45 & 43.78$\pm$3.24 & 52.43$\pm$3.51 & 51.89$\pm$3.59 & \first{72.62$\pm$0.82}  \\

\midrule


\multirow{4}{*}{Actor} 
&\emph{clean}          & \second{35.08$\pm$1.08}    & 29.25$\pm$1.21 & 24.36$\pm$1.11 & 30.72$\pm$0.72 & 30.03$\pm$0.13 & 34.37$\pm$0.69 & 33.69$\pm$0.73 & 30.02$\pm$0.91  & \first{35.91$\pm$0.92} \\
& Random               & \second{35.15$\pm$0.78}    & 27.59$\pm$1.12 & 25.64$\pm$1.02 & 30.16$\pm$1.09 & 28.36$\pm$1.95 & 25.41$\pm$0.72 & 34.19$\pm$0.59 & 28.92$\pm$1.03  & \first{35.29$\pm$0.82}  \\
& PRBCD                & \second{35.04$\pm$0.90}    & 27.76$\pm$1.66 & 24.95$\pm$0.89 & 30.48$\pm$1.28 & 28.37$\pm$1.95 & 27.21$\pm$0.64 & 26.23$\pm$0.79 & 28.66$\pm$2.01  & \first{35.49$\pm$1.13} \\
& Metattack            & \second{32.63$\pm$0.60}    & 28.00$\pm$1.10 & 25.54$\pm$0.75 & 30.34$\pm$1.04 & 28.45$\pm$1.26 & 28.29$\pm$0.42 & 26.97$\pm$0.65 & 29.65$\pm$1.12  & \first{33.02$\pm$0.10} \\
& Nettack              & \second{34.97$\pm$0.88}    & 28.87$\pm$0.73 & 25.51$\pm$0.95 & 30.86$\pm$0.96 & 28.60$\pm$1.20 & 25.96$\pm$0.86 & 27.20$\pm$0.74 & 30.05$\pm$0.81  & \first{35.36$\pm$1.31} \\
\bottomrule
\end{tabular}
\end{table*}

\begin{table*}[!htp]
\caption{White-box attack robust accuracy (\%) under 15\% perturbation for node classification.}
\label{tab:whitebox_results}
\centering
\fontsize{9pt}{10pt}\selectfont
\setlength{\tabcolsep}{2pt}
\begin{tabular}{c|cc|cccccc|c} 
\toprule
Dataset & Attack       & FROND               & Res-GRACE                & GraphACL          & PolyGCL           & LOHA             & EPAGCL  & SDMG         & FD-MVGCL    \\
\midrule
\multirow{3}{*}{Photo} 
&\emph{clean}          & 92.03$\pm$1.27     & 92.50$\pm$0.17            & 93.31$\pm$0.19    & 91.45$\pm$0.35    & 86.46$\pm$0.41   & 93.05$\pm$0.23   & \second{94.10$\pm$0.20}          & \first{94.27$\pm$0.85} \\
& PGD                  & 14.18$\pm$5.16     & 45.46$\pm$2.05            & 30.94$\pm$3.62    & 22.93$\pm$2.73    & \second{59.35$\pm$0.43}   & 5.02$\pm$2.85   & 11.98$\pm$0.81            & \first{92.58$\pm$0.64}         \\
& PRBCD                & 15.08$\pm$7.52     & 40.72$\pm$3.70            & 28.56$\pm$1.16    & 15.86$\pm$1.99    & \second{52.82$\pm$3.45}   & 9.91$\pm$1.73   & 8.71$\pm$6.22          & \first{79.24$\pm$0.84}      \\

\midrule
\multirow{3}{*}{Texas} 
&\emph{clean}          & \second{74.86$\pm$3.21}     & 54.59$\pm$5.51      & 71.08$\pm$0.34    & 72.51$\pm$1.95   & 67.75$\pm$2.70   & 68.92$\pm$4.05   & 53.60$\pm$2.67       & \first{80.19$\pm$0.81} \\
& PGD                  & \second{69.46$\pm$7.16}     & 28.85$\pm$8.55      & 21.62$\pm$7.20    & 57.84$\pm$14.05  & 47.30$\pm$16.45  & 13.78$\pm$6.78   & 30.49$\pm$4.32       & \first{76.49$\pm$4.84} \\
& PRBCD                & \second{65.14$\pm$3.91}     & 26.49$\pm$11.22     & 23.78$\pm$8.99    & 51.35$\pm$10.26  & 32.70$\pm$10.50  & 14.32$\pm$6.29   & 19.43$\pm$4.39       & \first{71.62$\pm$4.87} \\
\bottomrule
\end{tabular}
\end{table*}

\subsection{Robustness studies}\label{sec.robust_study}

We evaluate the robustness of FD-MVGCL through node classification under both black-box and white-box adversarial attacks on standard benchmarks, including homophilic (Cora, Citeseer, Photo) and heterophilic (Cornell, Actor, Texas) graphs. FD-MVGCL is compared against nine baselines: a robust supervised methods (FROND \cite{KanZhaDin:C24frond}), three robust GCL methods (GCL-Jac \cite{Kaidi2020}, Ariel \cite{shengyu2022}, Res-GRACE \cite{Lin2024ResGrace}), and five state-of-the-art GCL models (GraphACL, PolyGCL, LOHA, EPAGCL, SDMG).

\begin{table*}[!htb]
\caption{Comparison of training time (s), inference time (s), and GPU memory used (MiB) across different datasets.}
\label{tab:training_inference_storage}
\fontsize{9pt}{9pt}\selectfont
\setlength{\tabcolsep}{2.5pt}
\centering
\begin{tabular}{l|cccc|cccc|cccc}
\toprule
\multirow{2}{*}{Method} & \multicolumn{4}{c|}{Training Time (s)} & \multicolumn{4}{c|}{Inference Time (s)} & \multicolumn{4}{c}{GPU (MiB)} \\
                        & Cora & Wisconsin & Roman & Arxiv-year          & Cora & Wisconsin & Roman & Arxiv-year      & Cora & Wisconsin & Roman & Arxiv-year \\
\midrule
GraphACL                & 0.06 & 0.08   & 100.32  & 5403.97              & 0.004 & 0.01  & 0.15 & 1.06         & 932  & 1670   & 14008  & 53960 \\
PolyGCL                 & 0.34 & 0.23   & OOM     & OOM                  & 7.46  & 9.73   & OOM     & OOM             & 4098 & 894     & OOM   & OOM \\
EPAGCL                  &0.13  & 0.12   & 0.97    & OOM                  & 2.81   &  2.63  & 3.14   & OOM             & 649  &  329   & 22127  & OOM \\
GraphECL                & 0.02 & 0.27   & 60.58   & OOM                  & 0.50   & 0.04   & 0.04   & OOM             & 112  & 336   & 2306   &OOM  \\
SDMG                    & 0.06 & 0.54   & 0.36    & OOM                  & 1.22   & 0.28   & 0.55   & OOM             & 2888 & 824   & 3676   &OOM  \\
\midrule
FD-MVGCL                & 0.37 & 0.29   & 18.04   & 6.01                 & 1.47  & 1.41  &0.18 &  0.17           & 1330 & 906   & 16490  & 14824  \\
\bottomrule
\end{tabular}
\end{table*}

\subsubsection{Attack methods}
We consider four \emph{black-box} topology attacks in the evasion setting: Random, PRBCD \cite{Daniel2018}, Nettack \cite{Simon2021}, and Metattack \cite{zugner2019}. We also evaluate two \emph{white-box} attacks (i.e., PGD \cite{Madry2018} and PRBCD) that jointly perturb the graph structure and node features. All models are trained on clean graphs, with adversarial perturbations applied only during inference. The details of these attack methods are as follows:

\paragraph{Random attack} Randomly adds noisy edges by selecting node pairs across the graph. The number of added edges is controlled by a perturbation ratio relative to the original edge count.

\paragraph{PRBCD} Projected Randomized Block Coordinate Descent attack is a structure-based attack that perturbs the adjacency matrix $\bA$ by iteratively adding or removing edges to maximize the classification loss of a surrogate GNN (e.g., GCN). It uses a projected randomized block coordinate descent strategy under a fixed budget of edge modifications, enabling efficient and scalable adversarial perturbations. In the white-box setting, PRBCD further extends to jointly perturb node features by leveraging full access to model parameters and gradients.

\paragraph{Netttack} A targeted structure-based attack that perturbs the graph to mislead node classification. It estimates homophily and heterophily probabilities, removing edges to same-class nodes to reduce classification confidence and adding edges to different-class nodes to induce misclassification. Guided by a surrogate GNN, it greedily selects the most influential edges to modify within a given budget.

\paragraph{Metattack} A global structure-based attack that perturbs the adjacency matrix by computing meta-gradients of a surrogate GNN. It modifies the graph structure to maximize the overall classification loss, effectively degrading performance across the entire graph.

\paragraph{PGD} Projected Gradient Descent is a white-box attack that simultaneously perturbs the graph structure and node features to maximize the classification loss of the target model. It performs iterative gradient-based updates within a predefined perturbation budget, projecting the modifications back to the feasible space after each step. By leveraging full access to model parameters and gradients, PGD enables strong and precise attacks.

\subsubsection{Robustness results} We evaluate FD-MVGCL's robustness by varying the perturbation rate from $0\%$ to $25\%$ in $5\%$ increments, using random attacks on Photo (\cref{fig:photo_random_ratio}) and Metattack on Cora (\cref{fig:cora_meta_ratio}). FD-MVGCL exhibits the smallest performance degradation and consistently outperforms strong baselines such as GraphACL and PolyGCL. This robustness is attributed to the stability properties of fractional-order diffusion: smaller $\alpha_{k}$ values lead to reduced feature discrepancies as shown in \cref{eq.stab_fdgcl}, rendering the diffusion process more resistant to perturbations when $\alpha_{k} < 1$. The results in \cref{tab:blackbox_results,tab:whitebox_results} further confirm FD-MVGCL's superior robustness compared to baseline methods, particularly under more challenging white-box attack scenarios.

\subsection{Complexity analysis}
The training time complexity of FD-MVGCL consists of solving the FDE encoders and computing the contrastive loss. For a graph with $N$ nodes and $|\calE|$ edges, each FDE is solved over $E \coloneqq T / h$ time steps, where $h$ is the discretization step size and $T$ is the total integration time. At each step, the function $\calF\left(\bW, \bY_{j}\right)$ is evaluated and cached. Assuming $\mathcal{F}$ follows the GRAND model \cite{chamberlain2021grand}, each evaluation costs $C = |\calE| d$, where $d$ is the feature dimension. Solving one FDE costs $\calO(E C + E \log E)$, with $\log E$ term from fast convolution. For $\tilde{K}$ FDEs, the total cost is $\calO(\tilde{K}(E C + E \log E))$. The loss computation adds an additional cost $\calO(\tilde{K}N)$, resulting in a total training complexity of $\calO(\tilde{K}(E C + E \log E + N))$. 

Training/inference time and GPU memory (during training) comparisons with baselines (e.g., GraphACL, PolyGCL and EPAGCL) are shown in \cref{tab:training_inference_storage}. Although FD-MVGCL requires more GPU memory on large graphs (e.g., Roman and Ogbn-arxiv) due to maintaining intermediate states for multiple FDE solvers, it consistently avoids out-of-memory (OOM) issues and achieves lower overall training and inference costs compared with strong baselines. Moreover, apart from the orders $\alpha$, each diffusion module of FD-MVGCL introduces \emph{no other} learnable parameters, thereby contributing to its relatively low computational cost.

\subsection{Graph classification results}\label{graphclassification}

While most GCL methods target node-level representation learning and do not provide a straightforward graph-level extension, we assess FD-MVGCL's generality by applying a simple, non‑parametric readout (MeanPooling) to obtain graph-level embeddings. We evaluate this configuration on two standard graph-classification benchmarks, Proteins and DD, and compare against recent \textbf{graph-level} GCL models such as DRGCL and CI‑GCL as well as strong node-level baselines adapted to the graph-level setting (e.g., GraphACL). As reported in \cref{tab:graphresults}, FD‑MVGCL surpasses all node-level baselines including GraphACL and achieves results on par with \textbf{specialized} graph-level contrastive methods, demonstrating that the proposed fractional‑order multi‑view encoders generalize effectively beyond node-level tasks.


\begin{table}[!htb]
\caption{Graph classification results (\%). The top 3 methods are adapted from node-level GCL models, while the next 7 models are specialized graph-level GCL models.}
\label{tab:graphresults}
\centering
\setlength{\tabcolsep}{13pt}
\resizebox{0.48\textwidth}{!}{
\begin{tabular}{lcc}
\toprule
Method                                    & Proteins                        & DD \\
\midrule
MVGRL\cite{hassani2020mvgrl}             & 74.02$\pm$0.30                  & 75.20$\pm$0.40 \\
GraphACL\cite{Teng2023GraphACL}          & 73.50$\pm$0.70                  & - -        \\
SDMG \cite{zhu2025sdmg}                  & 73.16$\pm$0.16                  & 72.66$\pm$3.16          \\
\midrule
InfoGraph\cite{sun2019infograph}         & 74.44$\pm$0.40                  & 72.85$\pm$1.70 \\
GraphCL\cite{You2020GraphCL}             & 74.39$\pm$0.45                  & \second{78.62$\pm$0.40} \\
JOAO\cite{Youyoao2021}                   & 74.55$\pm$0.41                  & 77.32$\pm$0.54 \\
JOAO2\cite{Youyoao2021}                  & 75.35$\pm$0.09                  & 77.40$\pm$1.11 \\
SimGRACE\cite{XiaSimGrace2022}           & 75.35$\pm$0.09                  & 77.44$\pm$1.11 \\
DRGCL\cite{JiDRGCL2024}                  & 75.20$\pm$0.60                  & 78.40$\pm$0.70 \\
CI-GCL \cite{TanCIGCL2024}               & \first{76.50$\pm$0.10}          & \first{79.63$\pm$0.30} \\
\hline
FD-MVGCL                                 & \second{75.42$\pm$0.29}         & 78.45$\pm$0.43 \\
\bottomrule
\end{tabular}}
\end{table}

\section{Conclusion}
We have introduced FD-MVGCL, an adaptive multi-view, augmentation-free graph contrastive learning framework that leverages fractional-order graph neural diffusion to generate diverse, multi-scale representations. By treating the fractional order $\alpha$ as a learnable parameter, our approach dynamically adapts diffusion scales to the data, eliminating the need for handcrafted augmentations, negative sampling, or manual tuning. Extensive experiments demonstrate that FD-MVGCL achieves state-of-the-art performance across both homophilic and heterophilic benchmarks, while maintaining superior robustness under adversarial perturbations. This work establishes fractional-order dynamics as a principled and effective paradigm for multi-view contrastive learning on graphs. While FD-MVGCL achieves strong performance, it assumes a static graph structure. Future work will focus on extending our framework to dynamic graphs, enabling fractional diffusion to model evolving temporal and topological patterns.

\appendices

\section{Fractional-order derivatives and  GNNs}\label[Appendix]{appendix.fde_models}

The works \cite{KanZhaDin:C24frond,ZhaKanJi:C24} extend the ODE-based approaches by incorporating graph neural \gls{FDE}, generalizing the order of the derivative to positive real number $\alpha$. The motivation is the solution to an FDE that encodes historical information and thus retains the ``memory'' of the evolution. 

\subsection{Fractional-order derivatives}\label{sec.cap_diff}
Recall that the first-order derivative of a scalar function $f(t)$ is defined as the rate of change of $f$ at a time $t$:
\begin{align*}
    f'(t) = \frac{\ud f(t)}{\ud t}=\lim _{\Delta t \rightarrow 0} \frac{f(t+\Delta t)-f(t)}{\Delta t}
\end{align*}
It is local in the sense that $f'(t)$ is determined by the value of $f$ in a small neighborhood of $t$.

Fractional-order derivatives $D^{\alpha}_t$ generalize integer derivatives and the domain of $\alpha$ is extended to any positive real number. By the relation $D^{\alpha+1}_t= D^{\alpha}_tD^1_t$, it suffices to define $D^{\alpha}_t$ for the order parameter $\alpha\in (0,1)$. For the encoder design in this paper, we only consider $\alpha \in (0,1]$. The domain is large enough for us to choose encoders that generate different views. Moreover, solving a higher-order FDE requires additional information such as the initial derivative, which is usually not available.   

Although there are different approaches to defining $D^{\alpha}_t$, they all share the common characterization that the derivative is defined using an integral. Hence, the operator is ``global'' as historical values of the function are used. We show two approaches below.

The \emph{left fractional derivative} \cite{Sti23} of $f(t)$ is defined by the following integral
\begin{align*} 
    D^{\alpha}_tf(t) = \frac{1}{\Gamma(-\alpha)}\int_0^{\infty} \frac{f(t-\tau)-f(t)}{\tau^{1+\alpha}}\ud\tau.
\end{align*}

On the other hand, the \emph{Caputo fractional derivative} \cite{diethelm2010analysis} is defined as
\begin{align*} 
D_t^\alpha f(t)=\frac{1}{\Gamma(1-\alpha)} \int_0^t \frac{f^{\prime}(\tau)}{(t-\tau)^{\alpha}} \ud\tau.
\end{align*}
One can find other approaches and more discussions in \cite{tarasov2011fractional}.  

\subsection{Incorporating FDE}
To build a diffusion model based on FDE, it usually suffices to replace the integer order derivative with a fractional-order derivative in the governing differential equation. For example, the ODE in the GraphCON model \cite{rusch2022graph} is equivalent to a system of equations:
\begin{align*}
    \begin{aligned} 
    & \frac{\ud \bZ(t)}{\ud t} =\sigma\left(\mathbf{F}_\theta(\bY(t), t)\right)-\gamma \bY(t)-\nu \bZ(t) \\ 
    & \frac{\ud \bY(t)}{\ud t} =\bZ(t).\end{aligned}
\end{align*}
Replacing $\ud/\ud t$ with $D^{\alpha}_t$, one obtains its FDE version as 
\begin{align*}
    \begin{aligned} 
    & D_t^{\alpha} \bZ(t)=\sigma\left(\mathbf{F}_\theta(\bY(t), t)\right)-\gamma \bY(t)-\nu \bZ(t) \\ 
    & D_t^{\alpha} \bY(t)=\bZ(t).\end{aligned}
\end{align*}

A numerical solver is provided in \cite{KanZhaDin:C24frond}.

\section{Theoretical discussions} \label[Appendix]{sec:td}
\subsection{Formal version of \cref{thm:sgi-informal}}\label[Appendix]{sec.td_1}
In this section, we provide a rigorous discussion of the formal version of \cref{thm:sgi-informal}. To do this, we need the Mittag-Leffler functions, which were introduced by Mittag-Leffler in 1903 in the form of a Maclaurin series. We focus on a special case defined as follows:
\begin{align*}
e_{\alpha}(\lambda,t) = \sum_{n=0}^{\infty} (-1)^n\frac{\lambda^n t^{\alpha n}}{\Gamma(\alpha n + 1)}, \quad \lambda>0, \ t\geq 0,
\end{align*}
where $\Gamma(\cdot)$ is the Gamma function.

We make the following \emph{assumptions} regarding the encoder. The function $\calF$ in \cref{eq.frond_main} is the simple $-\overline{\bL}\bX$ as described after \cref{eq.ode}. For $D^{\alpha}_t$, we consider the left fractional derivative. For the model, we apply skip-connection after a diffusion of length $\tau$, and there are $m$-iterations so that the total time is $T=m\tau$. We shall consider $\alpha \in (0,1)$, and this a mild assumption, as we have seen that $D^1_t = \ud/\ud t$ is the limit of $D^{\alpha}_t$ when $\alpha \to 1$. The following formal version of \cref{thm:sgi-informal} analyzes the (asymptotic) properties of the Fourier coefficients of the output features. 

\begin{Theorem} \label{thm:sgi}
Suppose $\calG$ is a connected graph and $0<\alpha_l<\alpha_g\leq1$ are the order parameters. For subscript $s$ stands for either $l$ or $g$, let $n_s \geq 1$ be the integer such that $n_s\alpha_s < 1 \leq (n_s+1)\alpha_s$. Consider a graph signal $\bx$ with Fourier coefficients $\{c_i, 1\leq i \leq N\}$. Let $\bz_{\alpha_s}(T)$ be the output features of the encoders with order parameters $\alpha_s$. Its Fourier coefficients are $\{c_{\alpha_s,i}(T), 1\leq i\leq N\}$. Then the following holds:
\begin{enumerate}[(a)]
\item We have the expression: 
\begin{align*}
c_{\alpha_s,i}(T) = \parens*{ \sum_{j=0}^{n_s}b_{\alpha_s,i,j}\tau^{-j\alpha_s} + O\parens*{\ofrac{\tau}} } c_i,
\end{align*}
and $b_{\alpha_s,i,j}>0$. 
\item For fixed $l$ and $j$, the coefficient $b_{\alpha_s,i,j}$ is decreasing w.r.t.\ $i$.
\item For fixed $i$ and $j$, we have $b_{\alpha_l,i,j}> b_{\alpha_g,i,j}$. 
\end{enumerate}
\end{Theorem}

\begin{proof}
Let $\overline{\bL} = \bU\bLambda\bU\T$ be the orthogonal eigendecomposition of $\overline{\bL}$. As $\overline{\bL}$ is positive semi-definite, its (ordered) eigenvalues satisfy $0=\lambda_1 < \lambda_2 \leq \lambda_3 \ldots \leq \lambda_N$. Notice we have $\lambda_2>0$ as $\calG$ is connected. To solve the FDE $D^{\alpha}_t\bz_{\alpha}(t) = -\overline{\bL}\bz_{\alpha}(t)$, we perform eigendecomposition:
\begin{align*}
    D^{\alpha}_t\bz_{\alpha}(t) = -\bU\bLambda\bU\T\bz_{\alpha}(t),
\end{align*}
and hence
\begin{align*}
    D^{\alpha}_t\bU\T\bz_{\alpha}(t) = -\bLambda\bU\T\bz_{\alpha}(t). 
\end{align*}
Therefore, the spectral decomposition of the dynamics $\bz_{\alpha}(t)$ follows the equation $D^{\alpha}_t\bc_{\alpha}(t) = -\bLambda \bc_{\alpha}(t)$, where the $i$-th entry of $\bc_{\alpha}(t)$ is $c_{\alpha,i}(t)$. 

For consistency, we set $e_{\alpha}(0,t)=1$ the constant function independent of $\alpha$. Therefore, by \cite{Sti23}, we have the solution $c_{\alpha,i}(t) = e_{\alpha}(\lambda,t)c_{\alpha,i}(0), 1\leq i\leq N$, where $c_{\alpha,i}(0) = c_i$ depends only on the input feature. Recall we perform diffusion for a time $\tau$ followed by a skip-connection for $m$ iterations. The coefficients of the spectral decomposition of the final output features are 
\begin{equation} \label{eq:b1e}
\begin{split}
&c_{\alpha,i}(T) = \Big(1 +\ldots + e_{\alpha}(\lambda,\tau)\big(1+e_{\alpha}(\lambda,\tau)\big)\Big)c_{\alpha,i}(0) \\
= & \Big(1 + e_{\alpha}(\lambda,\tau) + \ldots + e_{\alpha}(\lambda,\tau)^m\Big)c_{\alpha,i}(0). 
\end{split}
\end{equation}
The case $c_{\alpha,i}(0)= c_i = 0$ is trivial as we can set $b_{\alpha_s,i,j}$ to be any positive number. We assume $c_{\alpha,i}(0) \neq 0$ for the rest of the proof, and want to estimate $1 + e_{\alpha}(\lambda,\tau) + \ldots + e_{\alpha}(\lambda,\tau)^m$ for $\alpha = \alpha_l$ and $\alpha_g$. 

By \cite{erd55}, if $\lambda_i>0$, $e_{\alpha_s}(\lambda_i, \tau)$ satisfies the following asymptotic estimation:
\begin{align} \label{eq:elt}
e_{\alpha_s}(\lambda_i, \tau) = \sum_{j=1}^{n_s}\frac{1}{\lambda_i^j\Gamma(1-j\alpha_s)}\tau^{-j\alpha_s} + O\parens*{\ofrac{\tau}}.
\end{align}

The Gamma function $\Gamma(\cdot)$ is positive and strictly decreasing on the interval $(0,1]$ (e.g., $\Gamma(1) = 1!=1$). Therefore, as $1-j\alpha_s >0$, the coefficient of $\tau^{-j\alpha_s}$ in \cref{eq:elt} is positive. Moreover, for $s$ denoting the subscripts $l$ and $g$ and $i\geq 2$, we have 
\begin{align} \label{eq:flg}
\begin{aligned}
    &\frac{1}{\lambda_i^j\Gamma(1-j\alpha_l)} > \frac{1}{\lambda_i^j\Gamma(1-j\alpha_g)} ~~\text{and}\\
     &\qquad \frac{1}{\lambda_i^j\Gamma(1-j\alpha_s)} \geq \frac{1}{\lambda_{i+1}^j\Gamma(1-j\alpha_s)}.
\end{aligned}
\end{align}
Now to compare the solution for the subscript $s$ being $l$ and $g$, we express $e_{\alpha_l}(\lambda_i,\tau)^k$ as follows
\begin{align*}
\begin{aligned}
e_{\alpha_l}(\lambda_i,\tau)^k &= \Big(\sum_{j=1}^{n_g}\frac{1}{\lambda_i^j\Gamma(1-j\alpha_l)}\tau^{-j\alpha_l}\\
&\qquad + \sum_{j=n_g+1}^{n_l}\frac{1}{\lambda_i^j\Gamma(1-j\alpha_l)}\tau^{-j\alpha_l} \Big)^k + O\parens*{\ofrac{\tau}}. 
\end{aligned}
\end{align*}
As compared with 
\begin{align*}
e_{\alpha_g}(\lambda_i,\tau)^k = \Big(\sum_{j=1}^{n_g}\frac{1}{\lambda_i^j\Gamma(1-j\alpha_g)}\tau^{-j\alpha_g} \Big)^k + O\parens*{\ofrac{\tau}},
\end{align*}
we see that up to $O(\frac{1}{\tau})$, $e_{\alpha_g}(\lambda_i,\tau)^k$ and hence $c_{\alpha_g,i}(T)/c_{\alpha_g,i}(0)$ (cf.\ \cref{eq:b1e}) consists of a summation of terms $\tau^{-j\alpha_g}$ for $1\leq j\leq n_g$, and the positive coefficient $\tau^{-j\alpha_g}$ is smaller than that of $\tau^{-j\alpha_l}$ in $c_{\alpha_l,i}(T)/c_{\alpha_l,i}(0)$. Moreover, $c_{\alpha_l,i}(T)/c_{\alpha_l,i}(0)$ has addition terms $\tau^{-j\alpha_l}, n_g<j\leq n_l$ with positive coefficients. This proves (a) and (c). From the second equality in \cref{eq:flg}, we obtain the conclusion that $b_{\alpha_s,i,j}$ is decreasing w.r.t.\ $i$, as claimed in (b).
 \end{proof}

To explain \cref{thm:sgi-informal}, $t$ in \cref{thm:sgi-informal} corresponds to $T = m\tau$ in \cref{thm:sgi}, and once the number of iterations $m$ is fixed, $\tau \to \infty$ if and only if $t\to \infty$. The signal $\bx$ in \cref{thm:sgi} is a component (i.e., a column) of the feature matrix $\bX$. By \cref{thm:sgi}, the Fourier coefficients $c_{\alpha_s,i}(t)$ of $\bY_{\alpha_s}(t)$, subscript $s$ being $l$ or $g$, relative to the Fourier coefficients of the input features, decrease as the frequency index $i$ increases. Moreover, the decrement for $\bY_{\alpha_l}(t)$ is much slower than that of $\bY_{\alpha_g}(t)$ as $t$ increases. The difference in the rate increases accordingly as $\alpha_g-\alpha_l$ becomes larger. As the leading Fourier coefficients ($i=0$) are both the same ($=m$), the Fourier coefficients of $\bY_{\alpha_l}(t)$ are more spread, whence the claims of \cref{thm:sgi-informal}. 

For most datasets studied in the paper, we choose $\alpha_l \leq 0.1$ and $\alpha_g> 0.5$, and thus $n_g=1$. By \cref{eq:b1e} and \cref{eq:elt}, the following holds for $c_{\alpha_g,i}(T)$:
\begin{align*}
c_{\alpha_g,i}(T)/c_{\alpha_g,i}(0) = 1 + \frac{1}{\lambda_i\Gamma(1-\alpha_g)}\tau^{-\alpha_g} + O\parens*{\ofrac{\tau}}.
\end{align*}
For $c_{\alpha_l,i}(T)/c_{\alpha_l,i}(0)$, we can identify a summand $1/\big(\lambda_i\Gamma(1-\alpha_l)\big)\cdot \tau^{-\alpha_g}$. The ratio between $1/\big(\lambda_i\Gamma(1-\alpha_l)\big)\cdot \tau^{-\alpha_l}$ and $1/\big(\lambda_i\Gamma(1-\alpha_g)\big)\cdot \tau^{-\alpha_g}$ is at least $\sqrt{\pi}/1.07 \cdot \tau^{0.4}$. This gives us some understanding of how these two sets of Fourier coefficients differ. 

\subsection{Stability}\label[Appendix]{sec:stability}


\subsubsection{Proof of \cref{theo:unified_perturbation}} 
By taking the difference between the original and perturbed dynamics, we obtain the following equation:
\begin{align*}
    D_t^\alpha(\bY(t) - \tilde{\bY}(t)) = \calF(\bW, \bY(t)) - \tilde\calF(\tilde{\bW},\tilde{\bY}(t)).
\end{align*}
Integrating both sides with the fractional integral operator $I_t^\alpha$ and then taking norms gives
\begin{align}
& \norm{\bY(t) - \tilde{\bY}(t)} \nn
&= \norm*{\bY(0) - \tilde{\bY}(0) + I_t^\alpha (\calF(\bW, \bY(t)) - \tilde\calF(\tilde{\bW},\tilde{\bY}(t))) }\nn
& \leq \norm{ \bY(0) - \tilde{\bY}(0)} \nn
&\qquad+ I_t^\alpha\norm{\calF(\bW, \bY(t)) - \tilde\calF(\tilde{\bW},\tilde{\bY}(t))}. \label{eq.diff_norm}
\end{align}
Using the triangle inequality, we bound the integrand of the second term on the right-hand side of \cref{eq.diff_norm} as follows:
\begin{align}
\label{eq.triangle}
\begin{aligned}
    &\norm{\calF(\bW,\bY(t))-\tilde\calF(\tilde\bW,\tilde\bY(t))}\\
    & \qquad\qquad\leq \norm{\calF(\bW,\bY(t)) - \calF(\bW,\tilde\bY(t))} \\
    & \qquad\qquad\qquad + \norm{\calF(\bW,\tilde\bY(t)) - \tilde\calF(\tilde\bW,\tilde\bY(t))}.
\end{aligned}
\end{align}
Given that $\calF$ is Lipschitz continuous with constant $L$, it follows that:
\begin{align}
\begin{aligned}
\label{eq.Lips_condi}
    &\norm{\calF(\bW,\bY(t)) - \calF(\bW,\tilde\bY(t))} \\
    &\qquad\qquad \qquad\leq L \norm{\bY(t)-\tilde\bY(t)}.
\end{aligned}
\end{align}
By definition of the combined perturbation $\varepsilon$, the second term on the right-hand side of  \cref{eq.triangle} is bounded as follows:
\begin{align}
\begin{aligned}
\label{eq.fun_per_bound}
  &\norm{\calF(\bW,\tilde\bY(t)) - \tilde\calF(\tilde\bW,\tilde\bY(t))}\\
  &\qquad\leq \sup_{\bY\in A} \norm{\calF(\bW,{\bY}) - \tilde\calF(\tilde{\bW},{\bY})} \leq \varepsilon.
\end{aligned}
\end{align}
Combining \cref{eq.diff_norm}, \cref{eq.Lips_condi} and \cref{eq.fun_per_bound}, we have the inequality
\begin{align}
\label{eq.comb_bound}
\begin{aligned}
    &\norm{\bY(t) - \tilde{\bY}(t)} \\
    &\qquad \leq LI_t^\alpha \norm{\bY(t)-\tilde\bY(t)}  +(1+ \frac{t^\alpha}{\Gamma(1+\alpha)})\varepsilon \\
    &\qquad \leq LI_t^\alpha \norm{\bY(t)-\tilde\bY(t)}  +(1+ \frac{T^\alpha}{\Gamma(1+\alpha)})\varepsilon,
\end{aligned}
\end{align}
where we have used the assumption that the initial perturbation satisfies $\norm{\bY_{0} - \tilde{\bY}_{0}} \leq \varepsilon$. 
Applying the fractional Gr\"onwall inequality \cite[Lemma 6.19.]{diethelm2010analysis} to \cref{eq.comb_bound}, we obtain 
\begin{align}
\label{eq.comb_bound_inequ}
    \norm{\bY(t) - \tilde{\bY}(t)} \leq \left(1+ \frac{T^\alpha}{\Gamma(1+\alpha)}\right)\varepsilon e_{\alpha}(-L,t^{\alpha}).
\end{align} 
Since the Mittag-Leffler function $e_{\alpha}(-L,t^{\alpha})$ can be bounded by $O\left(\frac{1}{1+LT^{\alpha}}\right)$, it follows from \cref{eq.comb_bound_inequ} that
\begin{align}
\begin{aligned}
    \norm{\bY(t)-\tilde{\bY}(t)} &\leq \left(1+ \frac{T^\alpha}{\Gamma(1+\alpha)}\right) \varepsilon O\left(\frac{1}{1+LT^{\alpha}}\right) \\
     &= O\left(\varepsilon T^{\alpha-1}\right)
\end{aligned}
\end{align}
for $t\in [0,T]$. 
This completes the proof of \cref{theo:unified_perturbation}.

\subsubsection{Proof of \cref{Theo.stab_fd_gcl}}
Subtracting the weighted average solution in \cref{eq.solu_fd_gcl} and the perturbation one yields
\begin{align}
\label{eq.diff_w_ave}
    \bY(t)-\Tilde{\bY}(t) = \sum_{k=1}^{\Tilde{K}}\beta_{k}\left(\bY_{k}(t)-\Tilde{\bY}_{k}(t)\right).
\end{align}
Take the norm of \cref{eq.diff_w_ave} and applying the triangle inequality, we obtain
\begin{align}
\norm{\bY(t)-\Tilde{\bY}(t)} \leq \sum_{k=1}^{\Tilde{K}}\beta_{k} \norm{\bY_{k}(t)-\Tilde{\bY}_{k}(t)}.
\end{align}
By applying \cref{theo:unified_perturbation} separately for each term $\norm{\bY_{k}(t)-\Tilde{\bY}_{k}(t)}$, we have
\begin{align}
 \norm{\bY_{k}(t)-\Tilde{\bY}_{k}(t)} = O\left(\varepsilon T^{\alpha_{k}-1}\right).
\end{align}
Thus, the final unified perturbation bound is given by:
\begin{align}
\begin{aligned}
    \norm{\bY(t)-\Tilde{\bY}(t)} &\leq \sum_{k=1}^{\Tilde{K}}\beta_{k} O\left(\varepsilon T^{\alpha_{k}-1}\right) \\
     & = O\left(\varepsilon T^{\alpha_{\Tilde{K}}-1}\right), 
\end{aligned}
\end{align}
for $t \in [0,T]$. This completes the proof of \cref{Theo.stab_fd_gcl}.

\section{Experimental details}\label{sec.implemen_details}
\subsection {Details of datasets}\label[Appendix]{dataset_statistic}
We refer the reader to \cref{tab:das} for detailed statistics of the datasets. 

\begin{table*}[!htp]
\caption{Statistics of homophilic and heterophilic graph datasets}
\centering
\setlength{\tabcolsep}{13pt}
\resizebox{0.8\textwidth}{!}{
\begin{tabular}{ccccccc}
\toprule
 Dataset             & Nodes      & Edges        & Classes  & Node Features & Data splits \\
\midrule
Cora                 &  2708      &  5429        &   7       &  1433        & standard          \\
Citeseer             &  3327      &  4732        &   6       &  3703        & standard          \\
PubMed               &  19717     &  88651       &   3       &  500         & standard          \\
Computer             &  13752     &  574418      &   10      &  767         & 10\%/10\%/80\%    \\  
Photo                &  7650      &  119081      &   8       &  745         & 10\%/10\%/80\%    \\
Ogbn-arxiv           &  169343    &  1166243     &  40       &  128         & standard          \\
\midrule
Texas                &  183       &  309         &   5       &  1793        & 48\%/32\%/20\%    \\
Cornel               &  183       &  295         &   5       &  1703        & 48\%/32\%/20\%    \\
Wisconsin            &  251       &  466         &   5       &  1703        & 48\%/32\%/20\%    \\
Squirrel-filtered    &  2205      & 46557        &   5       &  2089        & 48\%/32\%/20\%    \\
Chameleon-filtered   &  864       &  7754        &   5       &  2325        & 48\%/32\%/20\%    \\
Crocodile            &  11631     &  360040      &   5       &  2089        & 48\%/32\%/20\%    \\
Actor                &  7600      &  33391       &   5       &  932         & 48\%/32\%/20\%    \\
Roman-empire         &  22662     &  32927       &   18      &  300         & 50\%/25\%/25\%    \\
Amazon-ratings       &  24492     &  186100      &   5       &  300         & 50\%/25\%/25\%    \\
Arxiv-year           &  169343    &  1166243     &   5       &  128         & 50\%/25\%/25\%    \\
\bottomrule
\end{tabular}}

\label{tab:das}
\end{table*}

\subsection{Hyperparameter choices}\label[Appendix]{supp.hyperparameter}
We conduct all experiments on a machine equipped with an NVIDIA A5000 GPU with 24 GB of memory. A small grid search is performed to identify the optimal hyperparameters. Specifically, we search over the following ranges: 
diffusion depth $T\in \{1, 2, 3, 5, 10, 15, 20, 30\}$, step size $ h\in \{0.5, 1, 2, 3, 5, 10\}$, and hidden dimension $d\in \{256, 512, 1024, 2048, 4096, 8192\}$. The initial number of encoders is set to $K=5$, and threshold $\epsilon$ and $\delta$ in Algorithm 1 are both set to $0.0001$. In addition, the learning rate is selected from $\{0.001, 0.005, 0.01, 0.015, 0.2\}$, the weight decay from $\{0.0001, 0.0005, 0.00001\}$, and the regularization weight $\eta$ from $\{0.001,0.005,0.01,0.05,0.1,0.15,0.2,0.25,0.3\}$. The combination weight $\beta$ for aggregating $\bY_{k}$ is tuned over $[0,1]$ with step of $0.01$. 
The optimal configuration is chosen based on the average validation accuracy. A detailed summary of the selected hyperparameters and the final set of learnable parameters (e.g., $\tilde{K}$ and corresonding $\{\alpha_{k}\}_{k=1}^{\tilde{K}}$) for each dataset is provided in \cref{tab:hyper_parameter}. 

\begin{table*}[htb]
\caption{Details of the hyperparameters tuned by grid search on various datasets, along with the final set of learnable parameters.}
\centering
\fontsize{9pt}{10pt}\selectfont
\setlength{\tabcolsep}{4pt}
\begin{tabular}{lccccccccccc}
\toprule
Datasets    & $T$ & $h$     & $d$    &$\eta$ & $\text{lr}$  & $\text{weight decay}$ & $\text{epochs}$ & $\tilde{K}$ & $\{\alpha_{k}\}_{k=1}^{\tilde{K}}$          & $\{\beta_{k}\}_{k=1}^{\tilde{K}}$    \\
\midrule
Cora        & 20  & 1       & 512    & 0.05     & 0.01         & 0.0005               &  30              & 4       & $\{0.001, 0.529, 0.736, 1\}$               & $\{0.45, 0.02, 0.03, 0.5\}$             \\
Citeseer    & 5   & 0.5     & 2048   & 0.15     & 0.02         & 0.0005               &  20              & 5      & $\{0.009, 0.158, 0.405, 0.653, 1\}$         & $\{0.05, 0.05, 0.3, 0.05, 0.55\}$       \\
Pubmed      & 10  & 2       & 8192   & 0.1      & 0.01         & 0.0001               &  2               & 5      & $\{0.011, 0.259, 0.506, 0.752, 1\}$        & $\{0.01, 0.01, 0.08, 0.45, 0.45\}$       \\
Computer    & 5   & 0.5     & 4096   & 0.01     & 0.005        & 0.0005               &  2               & 5      & $\{0.001, 0.248, 0.515, 0.763, 1\}$         & $\{0.05, 0.05, 0.1, 0.65, 0.15\}$        \\
Photo       & 5   & 1       & 4096   & 0.05     & 0.005        & 0.0005               &  2               & 5      & $\{0.010, 0.258, 0.510, 0.990,1\}$         & $\{0.45, 0.02, 0.02, 0.01, 0.5\}$       \\
Ogbn-arxiv  & 30  & 3       & 256    & 0.2      & 0.01         & 0.0005               &  50              & 2      & $\{0.01, 1\}$                               & $\{0.2, 0.8\}$                         \\
\midrule
Squirrel    & 3   & 1       & 4096   & 0.05     & 0.01         & 0.0005               &  3               & 5      & $\{0.015, 0.561, 0.587, 0.706, 1\}$         & $\{0.5, 0.015, 0.015, 0.005, 0.465\}$         \\
Chameleon   & 3   & 1       & 2048   & 0.3      & 0.01         & 0.0005               &  2               & 5      & $\{0.012, 0.450, 0.641, 0.654, 1\}$         & $\{0.45, 0.01, 0.01, 0.03, 0.5\}$        \\
Crocodile   & 20  & 2       & 2048   & 0.01     & 0.01         & 0.0005               &  20              & 4      & $\{0.001, 0.589, 0.883, 1 \}$               & $\{0.4, 0.1, 0.1, 0.4\}$              \\
Actor       & 15  & 1       & 2048   & 0.01     & 0.01         & 0.0005               &  10              & 5      & $\{0.008, 0.354, 0.554, 0.827, 1\}$         & $\{0.7, 0.05, 0.05, 0.01,0.1\}$        \\
Wisconsin   & 20  & 2       & 2048   & 0.25     & 0.01         & 0.0005               &  30              & 5      & $\{0.001, 0.009, 0.113, 0.184, 1\}$         & $\{0.3, 0.5, 0.05, 0.05, 0.1\}$        \\
Cornell     & 30  & 5       & 2048   & 0.01     & 0.01         & 0.0005              &  50              & 4      &  $\{0.001, 0.018, 0.137, 1\}$                & $\{0.1, 0.6, 0.25, 0.05\}$                              \\
Texas       & 30  & 5       & 2048   & 0.01     & 0.01         & 0.0005              &  30              & 4      & $\{0.001, 0.063, 0.848, 1\}$                 & $\{0.1, 0.3, 0.05, 0.55\}$                          \\
Roman       & 2   & 1       & 8192   & 0.001    & 0.01         & 0.0005              &  2               & 2      & $\{0.01, 1\}$                                & $\{0.5,0.5\}$                    \\
Amazon      & 3   & 0.5     & 8192   & 0.001    & 0.001         & 0.0005             &  10              & 2      & $\{0.01, 1\}$                                & $\{0.99, 0.01\}$                     \\
Arxiv-year  & 2   & 1       & 1024   & 0.001    & 0.01          & 0.0005             &  2               & 5      & $\{0.008, 0.240, 0.523, 0.750,1\}$           & $\{0.025, 0.025, 0.075,0.025, 0.85\}$      \\
\bottomrule
\end{tabular}
\label{tab:hyper_parameter}
\end{table*}

\subsection{Contrastive loss functions}\label[Appendix]{supp.loss}
In scenarios where explicit negative samples are absent, non-contrastive methods instead optimize agreement between positive pairs. Representative approaches include knowledge-distillation frameworks such as BGRL \cite{thakoor2022BGRL}, which relies on cosine similarity, and redundancy-reduction techniques such as Barlow Twins \cite{zbontarbarlowtwin2021} and VICReg \cite{bardes2022vicreg}. To illustrate how different loss functions are formulated in this context, we consider two feature representations (or “views”), denoted by $\bY_{l}$ and $\bY_{g}$. Each consists of $N$ samples, where $\bY_{l,i}$ and $\bY_{g,i}$ represent the feature vectors of the $i$-th sample in the two views. Based on this setup, the Euclidean loss, Cosine similarity loss, Barlow Twins loss, VICReg loss and CCA loss are defined as follows.

\paragraph{Euclidean Loss} measures the squared $\ell_2$-norm between two feature representations $\bY_{l}$ and $\bY_{g}$, encouraging them to be as close as possible. It is defined as:
\begin{align*}
    \calL_{\text{Euclidean}} = \frac{1}{N} \sum_{i=1}^{N} \norm{\bY_{l,i}-\bY_{g,i}}_{2}^{2}.
\end{align*}

\paragraph{Cosmean Loss} measures the cosine similarity between two feature representations $\bY_{l}$ and $\bY_{g}$, encouraging their alignment. It is mathematically expressed as
\begin{align}
    \calL_{\text{Cosmean}} = 1 -\frac{1}{N} \sum_{i=1}^{N} \frac{\angles{\bY_{l,i},\bY_{g,i}}}{\norm{\bY_{l,i}}_{2}\norm{\bY_{g,i}}_{2}}
\label{cosmean_loss}
\end{align}
where $\angles{\bY_{l,i},\bY_{g,i}}$ is the inner product of these two vectors $\bY_{l,i}$ and $\bY_{g,i}$, and $\norm{\bY_{l,i}}_{2}$ and $\norm{\bY_{g,i}}_{2}$ are their respective $\ell_2$-norms. The cosmean loss is minimized when the feature vectors are perfectly aligned in the same direction, achieving maximum cosine similarity.

\paragraph{Barlow Twins Loss} \cite{zbontarbarlowtwin2021} is designed to encourage similarity between two feature representations $\bY_{l}$ and $\bY_{g}$, while reducing redundancy across dimensions within each representation. It is defined as:
\begin{align*}
    \calL_{\text{Barlow Twins}}=\sum_{i}\left(\left(1-\bC_{ii}\right)^{2}\right)+\lambda\sum_{i}\sum_{i\neq j} \bC_{ii}^2
\end{align*}
where $\bC = \frac{\bY_{l}^{\T}\bY_{g}}{N}$ is the cross-correlation matrix of the normalized feature representations $\bY_{l}$ and $\bY_{g}$, and $\lambda$ is a trade-off parameter. The first term minimizes the difference between the diagonal elements of $\bC$ and 1, ensuring the features from $\bY_{l}$ and $\bY_{g}$ are highly correlated. The second term minimizes the off-diagonal elements, promoting decorrelation between features and reducing redundancy. By balancing these two objectives, Barlow Twins Loss enables learning robust and diverse feature representations.

\paragraph{VICReg Loss} (Variance-Invariance-Covariance Regularization Loss) \cite{bardes2022vicreg} is designed to align two feature representations, $\bY_{l}$ and $\bY_{g}$, while ensuring variance preservation and minimizing redundancy. It consists of three components: variance regularization, invariance loss, and covariance regularization. The loss is expressed as:
\begin{align*}
    \calL_{\text{VICReg}} = \eta_1 \calL_{\text{inv}} + \eta_2 \calL_{\text{var}} + \eta_3 \calL_{\text{cov}},
\end{align*}
where $\eta_1$, $\eta_2$ and $\eta_3$ are hyper-parameters controlling the relative contributions of each term. The invariance component $\calL_{\text{inv}}$ is computed as the mean-squared Euclidean distance of the corresponding samples from $\bY_{l}$ and $\bY_{g}$:
\begin{align*}
    \calL_{\text{inv}} = \frac{1}{N} \sum_{i=1}^{N} \norm{\bY_{l,i}-\bY_{g,i}}_{2}^{2}.
\end{align*}
The variance component $\calL_{\text{var}}$ ensures that each feature dimension in $\bY_{l}$ and $\bY_{g}$ has sufficient variance to potentially prevent collapse. It is given by
\begin{align*}
\begin{aligned}
\calL_{\text{var}} = &\frac{1}{d} \sum_{j=1}^{d} \max(0,\varepsilon-\sqrt{\text{Var}(\bY_{l}[:,j])}) \\
    &\qquad\qquad + \max(0,\varepsilon-\sqrt{\text{Var}(\bY_{g}[:,j])})
\end{aligned}
\end{align*}
where $\bY_{l}[:,j]$ and $\bY_{g}[:,j]$ are the $j$-th feature columns of $\bY_{l}$ and $\bY_{g}$, respectively. $\text{Var}(\bY_{l}[:,j])=\frac{1}{N}\sum_{i=1}^{N}(\bY_{l}[i,j]-\mu_{l,j})^{2}$ is the variance of the $j$-th feature in $\bY_{l}$ (with $\mu_{l,j}=\frac{1}{N}\sum_{i=1}^{N}\bY_{l}[i,j]$), and $\varepsilon$ is a small positive constant to enforce nonzero variance. And the covariance regularization term $\calL_{\text{cov}}$ reduces redundancy by decorrelating different feature dimensions within each representation. It is expressed as:
\begin{align*}
    \calL_{\text{cov}} = \frac{1}{d}\sum_{i\neq j}\left(\text{Cov}(\bY_{l})_{j,k}^{2}+\text{Cov}(\bY_{g})_{j,k}^{2}\right)
\end{align*}
where $\text{Cov}(\bY_{l})_{j,k}=\frac{1}{N}\sum_{i=1}^{N}\left(\bY_{l,i,j}-\mu_{l,j}\right)\left(\bY_{l,i,k}-\mu_{l,k}\right)$ represents the off-diagonal elements of the covariance matrix for $\bY_{l}$ and likewise for $\bY_g$. The hyperparameters $\eta_1$, $\eta_2$, and $\eta_3$ control the relative contributions of variance regularization, invariance loss, and covariance regularization, respectively. By balancing these three terms, VICReg ensures robust feature alignment while maintaining diversity and decorrelation, making it well-suited for unsupervised learning tasks.

\paragraph{CCA Loss} (Canonical Correlation Analysis Loss) \cite{Zhang2021CCA_SSG} is self-supervised objective that balances invariance and decorrelation between two augmented views of the same data. Given two normalized feature matrices $\bY_{l}$ and $\bY_{g}$, the CCA-style learning loss is defined as
\begin{align*}
    \calL_{\text{CCA}} = \norm{\bY_{l}-\bY_{g}}_{F}^{2}+\lambda \left(\norm{\bY_{l}^{\T}\bY_{l}-\bI}_{F}^{2}+\norm{\bY_{g}^{\T}\bY_{g}-\bI}_{F}^{2}\right)
\end{align*}
where $\lambda$ is a non-negative trade-off parameter. The first (invariance) term encourages the two augmented views to produce similar representations, thereby maximizing cross-view correlation. The second (decorrelation) term regularizes each view to be approximately uncorrelated, ensuring that different feature dimensions encode diverse and complementary semantic information.

\begin{figure}[!htb]
    \centering
    \vspace{-3mm}
    \includegraphics[width=0.9\columnwidth]{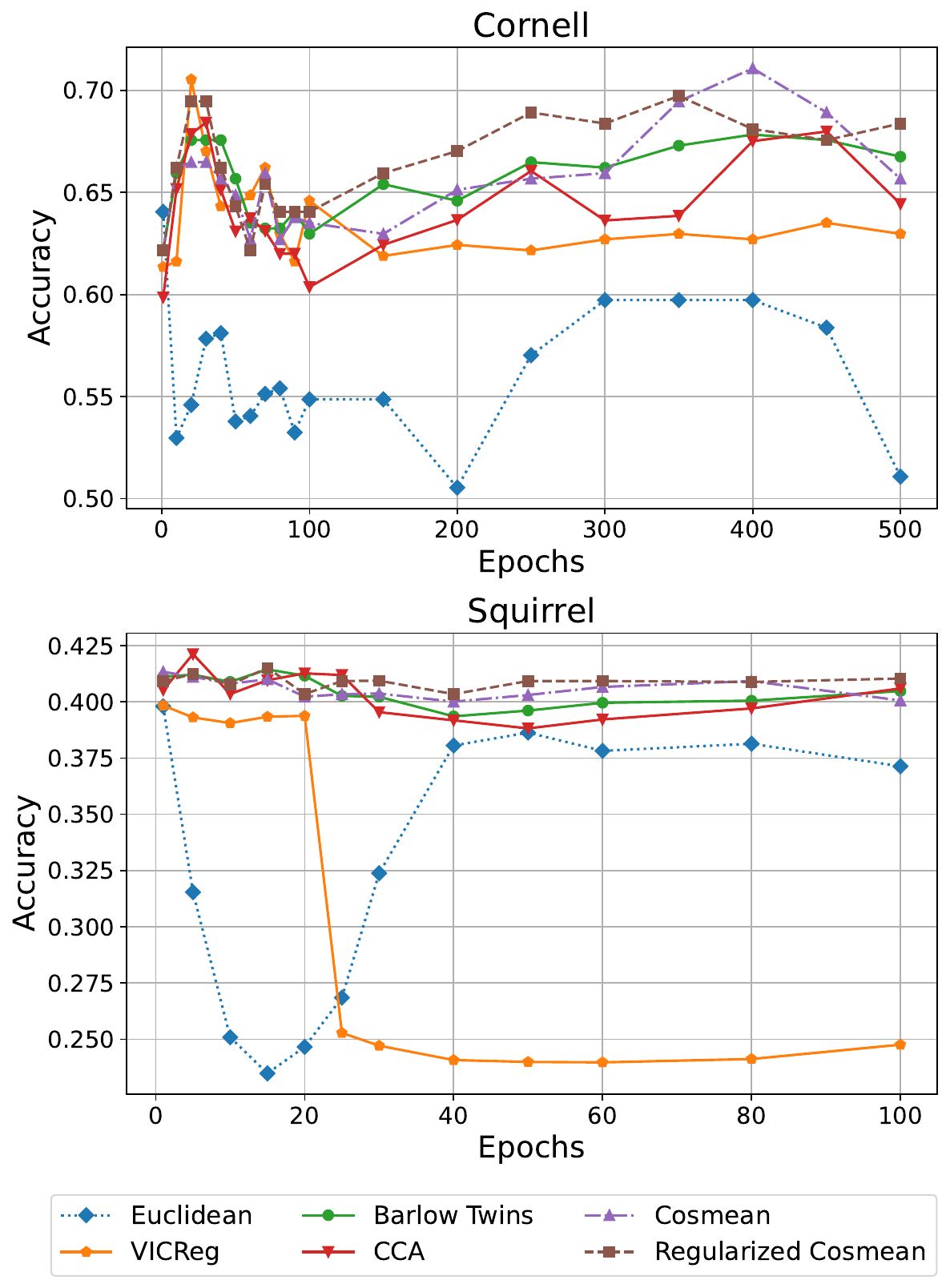}
    \centering
    \caption{Accuracy vs. training epochs for various loss functions on the Cornell and Squirrel datasets.}
    \label{fig.accuracy_loss_cornell_squirrel}
\end{figure}

\cref{fig.accuracy_loss_cornell_squirrel} presents additional classification accuracy results on two benchmark datasets: Cornell and Squirrel, evaluated above various contrastive loss functions. Consistent with the findings in the main text, these results further underscore the effectiveness of our proposed Regularized Cosmean loss. Unlike other loss functions, which tend to experience performance degradation as training progresses, the Regularized Cosmean loss demonstrates superior stability by maintaining consistent accuracy across epochs. These results provide additional evidence of its ability to mitigate dimension collapse and ensure robust performance. 

\section{T-SNE visualizations of node features}\label[Appendix]{supp.tsne}

\begin{figure}[!htbp]
    \centering
    \includegraphics[width=1\linewidth]{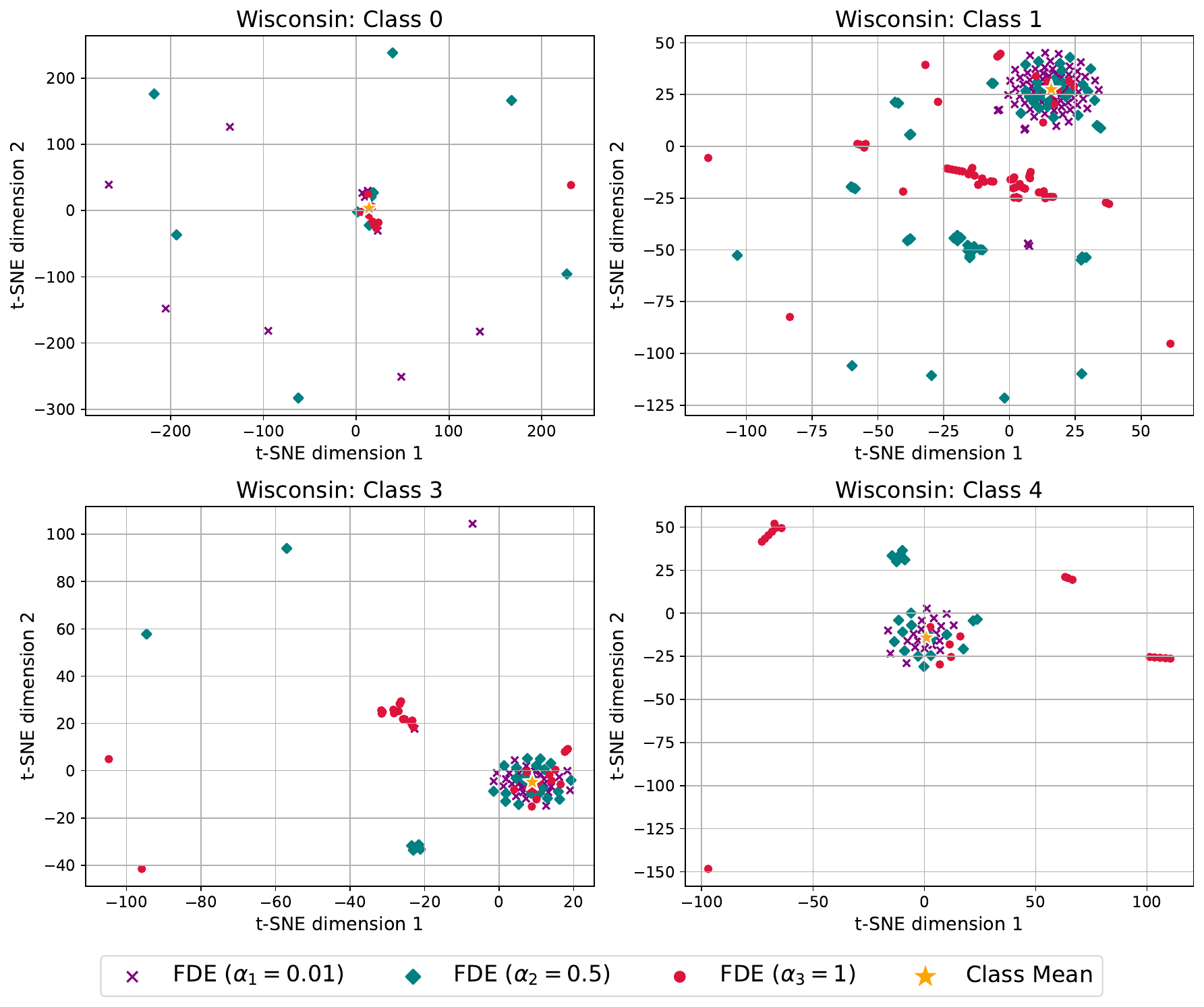}
    \caption{t-SNE visualizations of Wisconsin node features for different class labels.}
    \label{fig:tsne_wisconsin_horizontal}
\end{figure}
\begin{figure}[!htbp]
    \centering
    \includegraphics[width=1\linewidth]{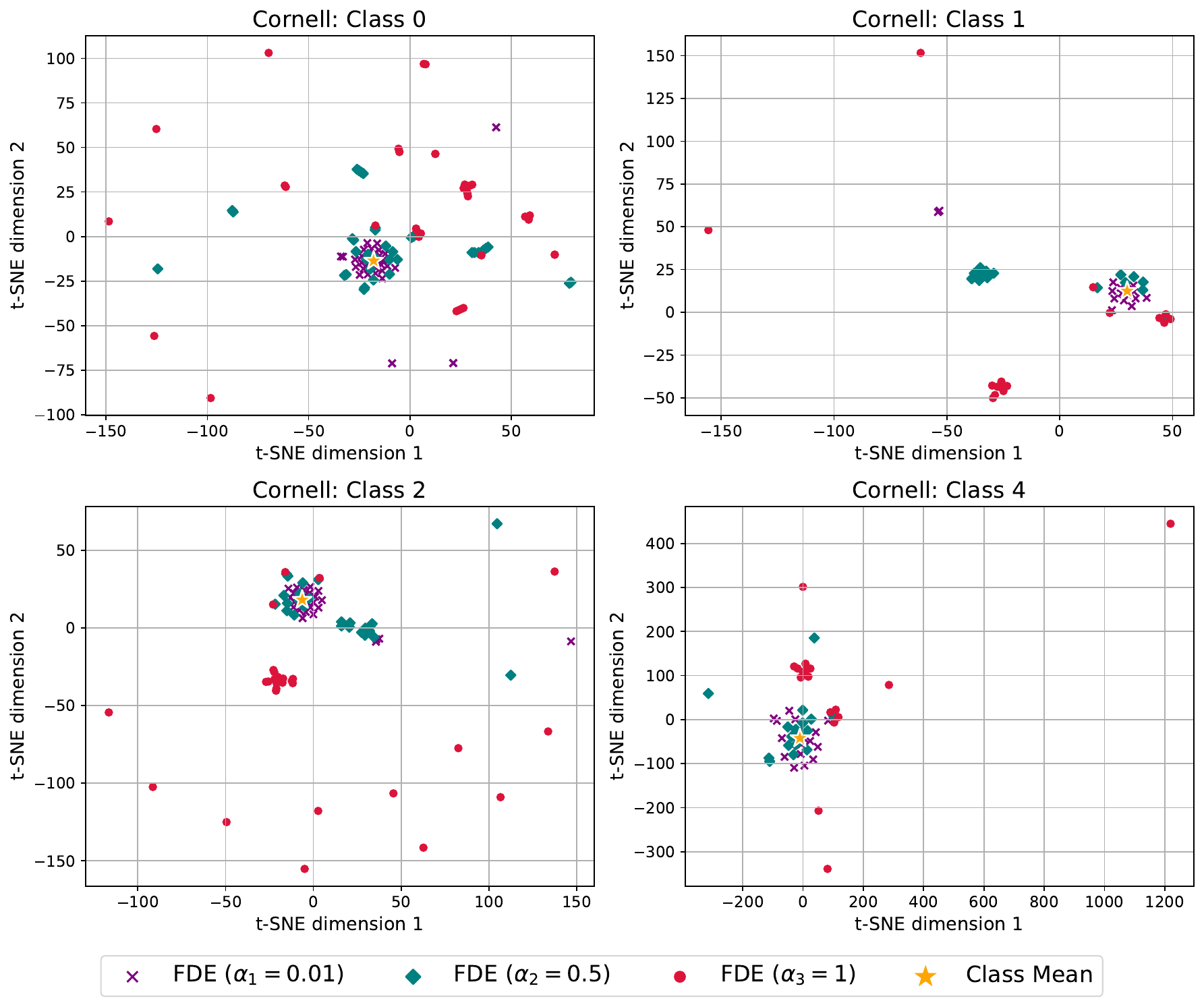}
    \caption{t-SNE visualizations of Cornell node features for different class labels.}
    \label{fig:tsne_cornell_horizontal}
\end{figure}
\begin{figure*}[!htbp]
    \centering
    \includegraphics[width=\linewidth]{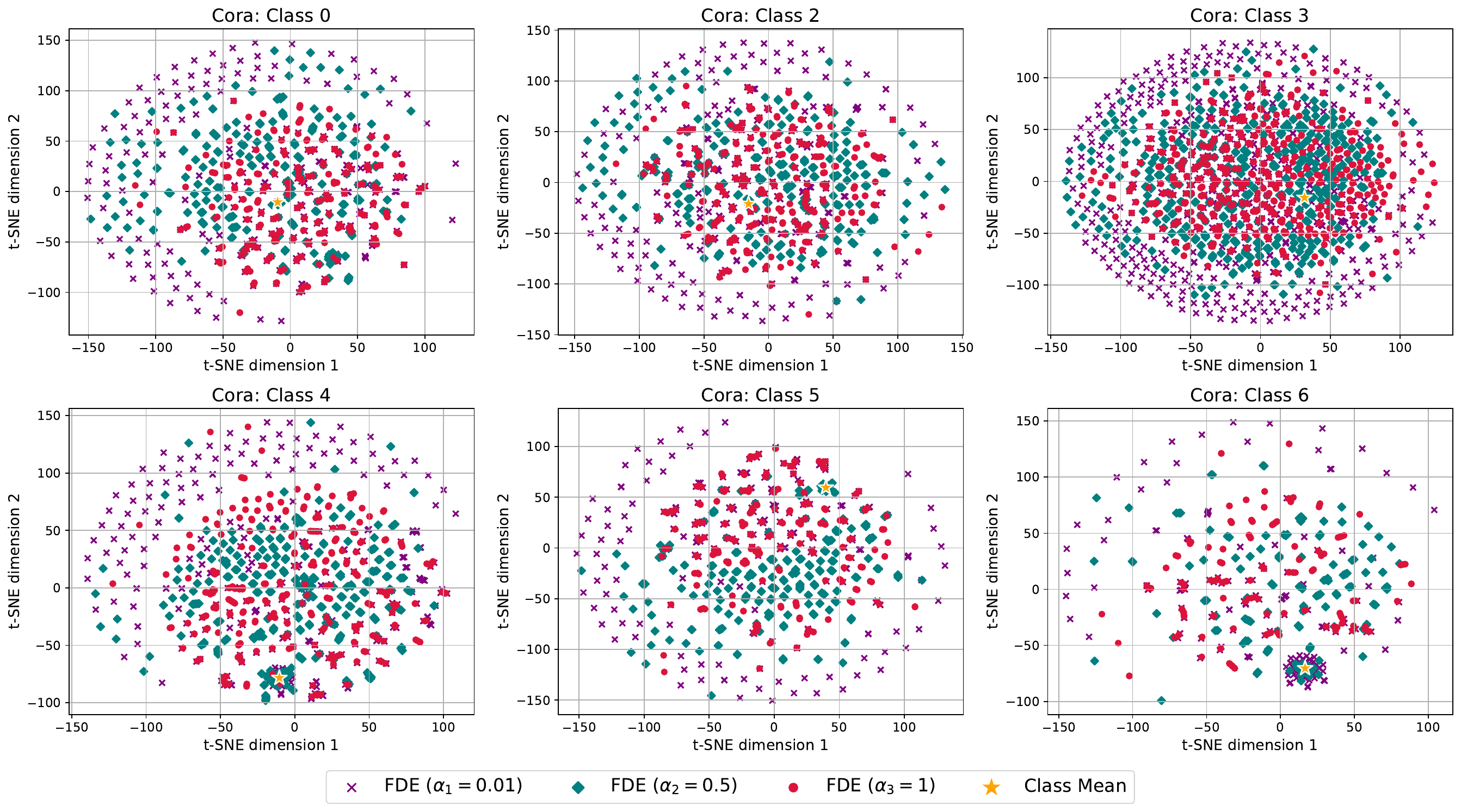}
    \vspace{-5mm}
    \caption{t-SNE visualizations of node features from selected Cora labels. Each pair shows embeddings from FDE encoders with different $\alpha$.}
    \label{fig:tsne_cora_all}
\end{figure*}

We present additional t-SNE visualizations of node features for each class in the Cora (homophilic), Wisconsin and Cornell (heterophilic) datasets, shown in \cref{fig:tsne_wisconsin_horizontal}, \cref{fig:tsne_cornell_horizontal} and \cref{fig:tsne_cora_all} respectively. These plots are generated using encoders with varying fractional-order parameters, highlighting the distinct embedding patterns produced by different diffusion dynamics.

\end{document}